\documentclass{article}

\PassOptionsToPackage{numbers,compress}{natbib}

\usepackage[eandd, preprint]{arxiv_neurips}

\usepackage[utf8]{inputenc} 
\usepackage[T1]{fontenc}    
\usepackage{hyperref}       
\usepackage{url}            
\usepackage{booktabs}       
\usepackage{amsfonts}       
\usepackage{nicefrac}       
\usepackage{microtype}      
\usepackage{xcolor}         
\usepackage{booktabs}
\usepackage{multirow}
\usepackage{makecell}
\usepackage{graphicx}   
\usepackage{adjustbox}
\usepackage{subcaption}
\usepackage{float}
\long\def\zp#1{#1}
\long\def\gc#1{#1}

\title{ImageAttributionBench: How Far Are We from Generalizable Attribution?}

\author{%
  Tingshu Mou$^{1}$, Zhipeng Wei$^{2}$, Chao Gong$^{1}$, Jingjing Chen$^{1\dagger}$, Xingjun Ma$^{1\dagger}$
  \\
  $^{1}$Fudan University, $^{2}$University of California, Berkeley
}
\begin{document}

\maketitle

\begingroup
\renewcommand{\thefootnote}{}
\footnotetext{$^{\dagger}$Corresponding author.}
\endgroup

\begin{abstract}
The rapid advancement of generative AI has enabled the creation of highly realistic and diverse synthetic images, posing critical challenges for image provenance and misinformation detection. This underscores the urgent need for effective image attribution. However, existing attribution datasets are constrained by limited scale, outdated generation methods, and insufficient semantic diversity—hindering the development of robust and generalizable attribution models. To address these limitations, we introduce \textbf{ImageAttributionBench}, a comprehensive dataset comprising images synthesized by a wide array of advanced generative models with state-of-the-art (SOTA) architectures. Covering multiple real-world semantic domains, the dataset offers rich diversity and scale to support and accelerate progress in image attribution research. To simulate real-world attribution scenarios, we evaluate several SOTA attribution methods on ImageAttributionBench under two challenging settings: (1) training on a standard balanced split and testing on degraded images, and (2) training and testing on semantically disjoint splits. In both cases, current methods exhibit consistently poor performance, revealing significant limitations in their robustness and generalization to unseen semantic content. Our work provides a rigorous benchmark to facilitate the development and evaluation of future image attribution methods.
\end{abstract}


\begin{figure*}[t]
  \centering
  \includegraphics[width=1\textwidth]{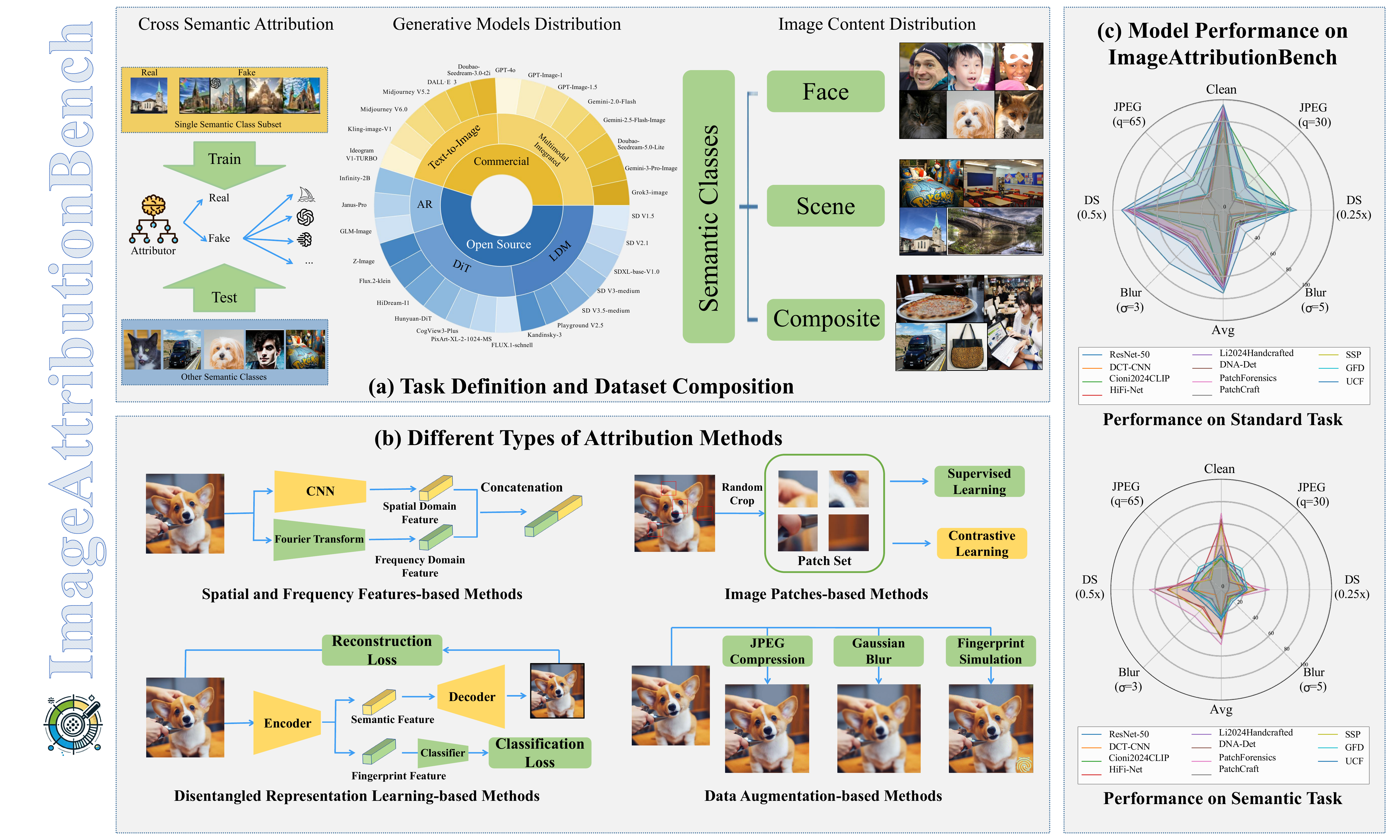}
  \caption{Overview of ImageAttributionBench. (a) The cross-semantic attribution task and key statistics of the constructed dataset. (b) Illustration of the four main categories of existing attribution methods. (c) Performance comparison of representative attributors on standard and semantic tasks under various degradations.}
  \label{fig:head}
\end{figure*}

\section{Introduction}
\label{sec:intro}

\zp{
In recent years, the emergence and evolution of auto-regressive models~\cite{esser2021taming, li2024autoregressive, tian2024visual} and diffusion models~\cite{ho2020denoising,  song2020denoising, dhariwal2021diffusion, nichol2021glide, rombach2022high, podell2023sdxl, ramesh2022hierarchical} have led to AI-generated content (AIGC) becoming increasingly realistic and widely applied across industries, bringing convenience to fields such as entertainment~\cite{sora,aidungeon,sudowrite}, advertising~\cite{kietzmann2018artificial, du2023effect}, and medicine~\cite{shao2024artificial, zhang2023generative}.
This progress is particularly evident in AI-synthesized images, which have seen gradual improvements in resolution and semantic consistency, accompanied by more accessible generation methods for users.
}

However, issues such as the spread of misinformation~\cite{xu2023combating, sophia2025social, yu2024fake}, privacy violations~\cite{manheim2019artificial,  jin2018artificial}, and fraud~\cite{guo2023aigc, wang2024study} have become more serious. Therefore, the importance and necessity of research on traceability technologies to track the origin of synthesized images have steadily increased. \zp{This urgency has brought }image attribution, also known as model attribution, \zp{to the forefront as an essential task}. It involves identifying the source model of an AI-generated image—that is, determining which generative model was used to create the image.

\zp{In response to this crucial task of image attribution, several datasets have been proposed and developed over the past years to facilitate research and evaluation~\cite{yu2019attributing, wang2020cnn, guo2023hierarchical, bui2022repmix, yang2023progressive, zhu2023genimage}.
Early image attribution datasets~\cite{yu2019attributing, wang2020cnn} primarily use low-resolution images from single-category GAN models, covering limited GAN architectures and specific semantic domains like faces and bedrooms. Subsequent datasets~\cite{bui2022repmix, yang2023progressive, guo2023hierarchical, zhu2023genimage} expand the scale and diversity of models and images by including various GAN types, semantic categories, and the introduction of diffusion models.

Despite these advancements, existing datasets suffer from critical limitations that hinder the development of robust and generalizable attribution methods:
(1) \textbf{Limited Model Diversity:} The restricted number of generative models (typically $\leq$15) within these datasets inadequately reflects the complexity of real-world scenarios. 
(2) \textbf{Reliance on Outdated Generation Methods:} These datasets predominantly feature older GAN architectures and early diffusion models, lacking contemporary cutting-edge paradigms like Diffusion Transformers (DiT) and Auto Regressive (AR) models. 
(3) \textbf{Insufficient Semantic Diversity:} The limited range of semantic categories in these datasets can lead to entanglement between model-specific fingerprints and semantic features, thereby compromising the generalization capabilities of attribution models.
}

\zp{
To address the critical limitations of existing datasets, we introduce ImageAttributionBench, a novel benchmark designed to advance the field. To better reflect the complexities of real-world image generation, ImageAttributionBench features a comprehensive suite of 31 state-of-the-art generators, comprising 17 open-sourced and 14 commercial models. Significantly, beyond conventional Latent Diffusion Models (LDM) \cite{rombach2022high}, our dataset uniquely integrates cutting-edge architectures that have gained prominence in the last two years, including DiT, AR, and native multi-modal generative models. 
Furthermore, ImageAttributionBench introduces semantic diversity by including 10 semantic classes sourced from 6 distinct real-world datasets, thus representing a broader range of domains than previous benchmarks.
ImageAttributionBench comprises roughly 640,000 images, collected from real-world datasets and synthesized using this diverse set of 31 generative models.
}

\zp{Table~\ref{table-datasets} provides a comparative analysis of ImageAttributionBench against existing datasets. A key design decision in our work is the exclusion of the somewhat outdated GAN architecture. It allows us to concentrate on a more extensive and varied set of advanced generative models that represent the current state of the art.
}
\zp{Besides, the design of separate classes within datasets encourages attribution methods to capture generative-model-specific features that are independent of semantics.
}

\zp{
To demonstrate the inherent challenges presented by ImageAttributionBench, we conduct comprehensive experiments using state-of-the-art methods from both fake image detection and attribution literature. Our evaluation encompasses two distinct settings: a standard balanced split and a more rigorous semantic-based split specifically designed to assess the generalization capabilities of these methods across diverse semantic categories.
Figure~\ref{fig:head} illustrates the overall performance of the evaluated methods on ImageAttributionBench.
While all attribution methods achieve high accuracy on the standard balanced split where all semantic classes are available, their classification performance declined sharply when access is restricted to a single semantic class. 
This stark contrast underscores the crucial role of semantics in robust image attribution, highlighting the potential of ImageAttributionBench to drive the development of more realistic and effective attribution techniques.
} 
Our code and dataset are available at \url{https://github.com/mttry/ImageAttributionBench}.
\begin{table*}  
 \caption{Overview of existing attribution datasets.
Size indicates the number of images.
Models lists only neural-network generative models.
Semantic Diversity indicates whether sufficient semantic classes are included.
Semantic Balance indicates whether semantic distribution is balanced across generative 
 models.} 
  \label{table-datasets}  
  \centering  
   \resizebox{\textwidth}{!}{
\begin{tabular}{llllll}
\toprule
Dataset & Size & \makecell[c]{Models \\ (by Type and Quantity)} & \makecell[c]{Semantic \\ Diversity} & \makecell[c]{Semantic \\ Balance} \\
\midrule

\citet{yu2019attributing} 
& 550k & 4 × GAN & No & Yes \\

\gc{ForenSynths~\cite{wang2020cnn}}
& 72k & 6 × GAN & Yes & No \\

Attribution88~\cite{bui2022repmix} 
& 1M & 7 × GAN & Yes & Yes \\

IFDL~\cite{guo2023hierarchical} 
& 1.9M & 4 × GAN + 4 × Diffusion & Yes & No \\

OSMA~\cite{yang2023progressive} 
& 300k & 4 × GAN & Yes & No \\

GenImage~\cite{zhu2023genimage} 
& 2.7M & 1 × GAN + 7 × Diffusion & Yes & Yes \\

WildFake~\cite{hong2024wildfake}
& 3.69M & 9 × GAN + 8 × Diffusion + 4 × Other & Yes & No \\

WILD~\cite{bongini2025wild}
& 50k & 8 × Commercial + 3 × Diffusion + 4 × DiT + 5 × Others & No & No \\

\textbf{ours} 
& 640k & \textbf{14 × Commercial + 7 × Diffusion + 7 × DiT + 3 × AR} & \textbf{Yes} & \textbf{Yes} \\
\bottomrule
\end{tabular}
  }
\end{table*}

\section{Related Work}
\label{sec:related_work}
\label{related work}
\zp{
We categorize current image attribution methods into four key areas: spatial and frequency features, data augmentation, image patches, and disentangled representation learning. 
We present these methods below.
}

\paragraph{Spatial and Frequency Features-based Methods} These methods exploit image statistics and high-frequency features introduced during generations for image attribution.  
\citet{frank2020leveraging} pioneer early work by identifying model-specific fingerprints based on Discrete Cosine Transform (DCT) spectrum artifacts.  
More recently, \citet{guo2023hierarchical} propose HiFi-Net, a multi-branch architecture operating at different resolutions to fuse features from both color and high-frequency domains. HiFi-Net incorporates a localization module, enabling it to detect manipulated regions in CNN-synthesized or edited images.  
Leveraging powerful pretrained models, \citet{cioni2024clip} demonstrate that features extracted using CLIP, combined with a simple linear classifier, achieve state-of-the-art performance on the challenging open-world source attribution task.  
Building on this, \citet{li2024handcrafted} \zp{combine CLIP embeddings with traditional handcrafted filter-extracted features to improve performance, particularly in small-sample training datasets.}

\paragraph{Data Augmentation-based Methods} These methods improve the generalization and robustness of the attributor by enhancing the training data.  
\citet{yang2022deepfake} propose DNA-Det, leveraging data augmentation through pretraining on image transformation classification combined with patch-wise contrastive learning to achieve architectural-level attribution. 
Subsequently, \citet{bui2022repmix} \zp{propose a feature mixing-based augmentation strategy.}
Building on the idea of DNA-Det, \citet{yang2023progressive} introduce POSE, which simulates model fingerprints using lightweight augmentation models composed of two convolutional layers.

\paragraph{Image Patches-based Methods} These methods claim that the fingerprint traces left by models can be detected in small regions, allowing efficient attribution using only image patches. 
\citet{chai2020makes} pioneer patch-based analysis in fake image detection, using shallow networks with limited receptive fields to localize transferable local artifacts. 
In more recent studies, PatchCraft~\cite{zhong2023patchcraft} posits that artifacts in texture-rich and texture-simple areas of generated images are inconsistent, unlike authentic images. 
\zp{Its approach involves segmenting images into patches and then reassembling them based on their texture complexity to facilitate fingerprint extraction.}
Furthermore, SSP \cite{chen2024single} uses texture-simple patches, which contain rich information for fingerprint extraction, because they are often neglected by generative models compared to texture-rich patches.

\paragraph{Disentangled Representation Learning-based Methods}  These methods aim to separate the forgery traces of the image from the semantic content. GFD \cite{yang2021learning} uses an encoder-decoder and an auxiliary classifier to classify real images with added fingerprints as forged, promoting  fingerprint disentanglement. Building upon GFD, UCF \cite{yan2023ucf} separately extracts semantic and forged features, reconstructs images from semantic features for disentanglement, using only forged  for attribution. 

\section{Dataset Construction}

\begin{figure*}[t]
  \centering
  \includegraphics[width=1\textwidth]{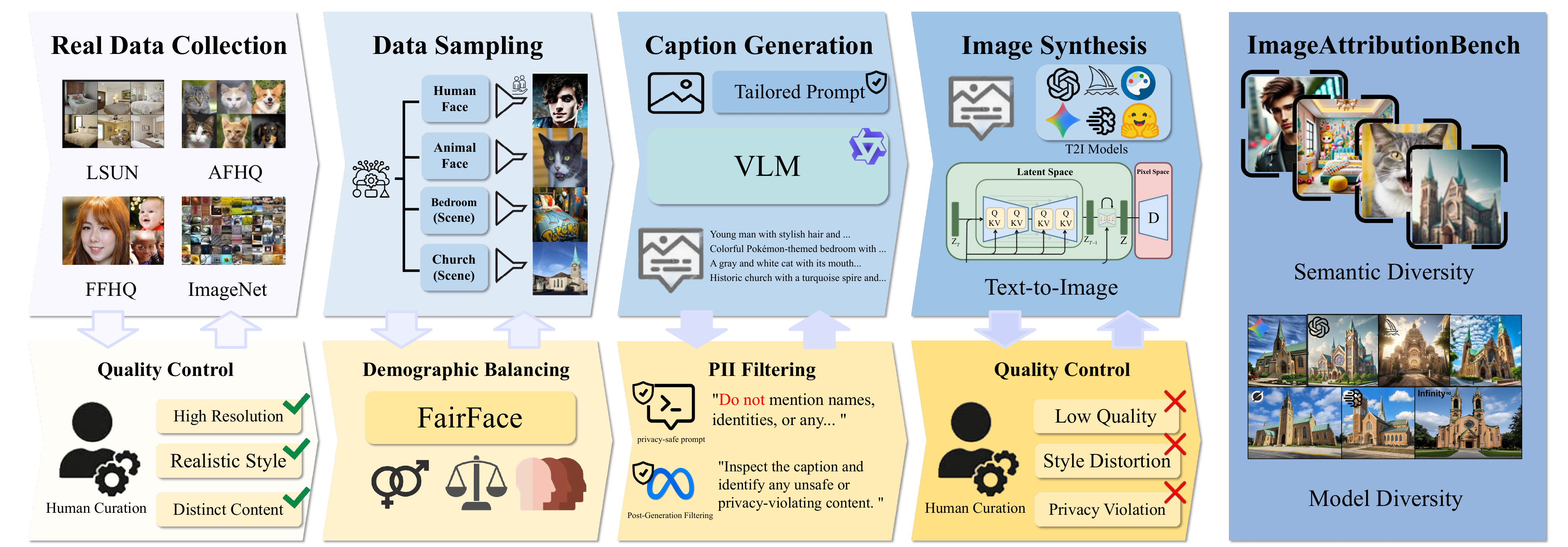}
  \caption{Pipeline of our dataset construction.}
  \label{fig:pipline}
\end{figure*}

\zp{
The rapid advancement of generative AI necessitates a paradigm shift in image attribution research, yet existing datasets suffer from limitations in scale, outdated generation methodologies, and a critical lack of semantic diversity. To bridge this significant gap, we present ImageAttributionBench. Our dataset is designed to provide both the necessary large-scale data and the crucial semantic breadth required to effectively evaluate and advance the field.
Specifically, 
}
the construction of our ImageAttributionBench dataset comprises three key steps: (1) collecting and sampling diverse real-world images from real datasets, ensuring broad coverage of visual concepts; (2) leveraging state-of-the-art large vision-language models to generate accurate and descriptive captions for these images; and (3) employing cutting-edge text-to-image models to synthesize images guided by these captions. The complete dataset construction pipeline is illustrated in Figure~\ref{fig:pipline}.

For each step, we employ human and AI-assisted curation to guarantee the quality and ethical integrity of our dataset. Specifically, we employ human curation during real data collection and image generation to maintain high visual quality in both real and synthesized samples. To further eliminate potential ethical concerns, we apply demographic balancing during data sampling and conduct Personally Identifiable Information (PII) filtering during caption-generation. Details of the implementation are provided in Section~\ref{ethic} of the supplemental material.

\subsection{Real Data Collection}
\zp{
To ensure our ImageAttributionBench benefits from a broad spectrum of real-world visual concepts and semantic diversity, we initiated the dataset construction by selecting several domain-specific image datasets. 
These include CelebA-HQ~\cite{karras2017progressive} and FFHQ~\cite{karras2019style} for human faces, AFHQ~\cite{choi2020stargan} for animal faces, LSUN~\cite{yu15lsun} for scenes, and more general ImageNet-1k~\cite{ILSVRC15} and COCO~\cite{lin2014microsoft}.
}
Specifically, for CelebA-HQ and FFHQ, we sample 2,000 images each to cover two classes of human faces. For AFHQ and LSUN, we sample 6,000 images each to cover dog, cat, wild, church, classroom, and bedroom. For ImageNet-1k, which has 1,000 labels, we sample 2 images per label, totaling 2,000 images. For COCO, we randomly sample 2,000 images. 
\zp{The combination of these sampling efforts results in a real-world image subset of 20,000 images spanning 10 semantic classes.}

\begin{figure*}[t]
  \centering
  \includegraphics[width=1\textwidth]{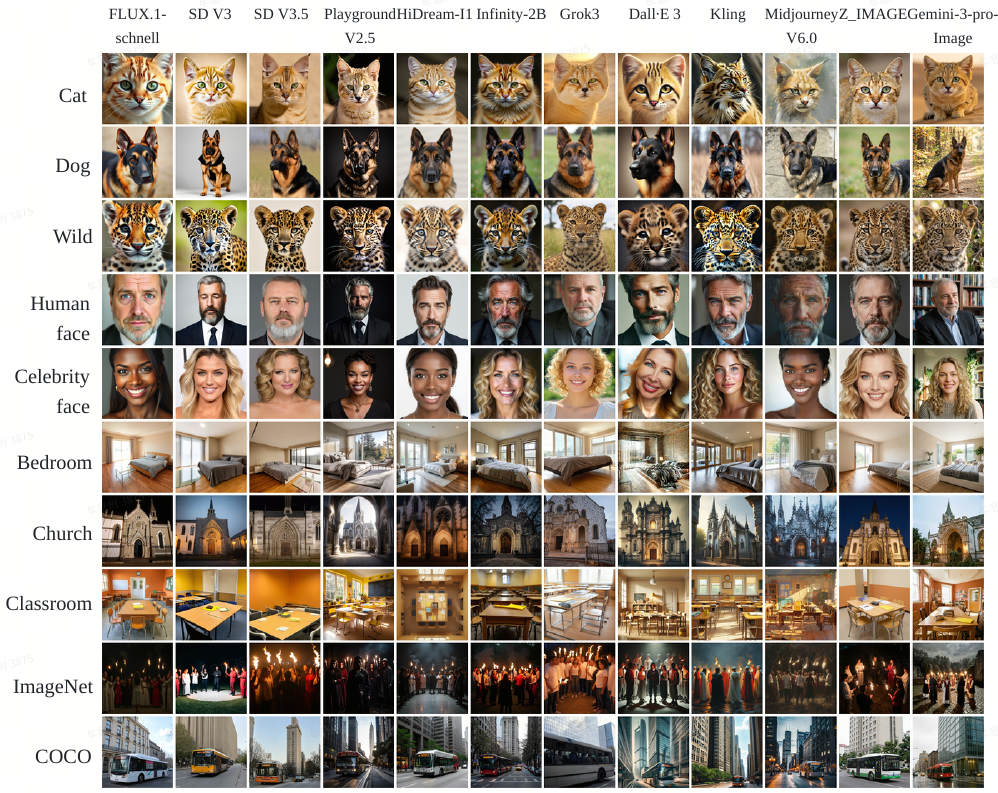}
  \caption{Visualization of images in ImageAttributionBench.}
  \label{fig:sample}
\end{figure*}

\begin{table}  
  \caption{Semantic classes and their corresponding prompt templates.}  
  \label{table-prompts}  
  \centering  
  \resizebox{\linewidth}{!}{
  \begin{tabular}{lp{10cm}}  
    \toprule  
    Semantic Class & Prompt Template \\
    \midrule  
    Human Face & Describe the face in the image by focusing only on general facial features. Include aspects such as the shape and appearance of the eyes, nose, mouth, and ears, the color and texture of the skin or fur, facial expression, and any visible hair or fur. Do not mention names, identities, or any personally identifiable information. The description should clearly center on the face, with minimal reference to the background or body. Keep it vivid and concise, 20 to 50 words.
  \\
    Animal Face & Describe the face in the image with focus on facial features.  Include details such as the shape and appearance of the eyes, nose, mouth, and ears, the color and texture of the skin or fur, facial expression, and any visible hair or fur.  The description should clearly center on the face, with minimal reference to the background or body.  Keep it vivid and concise, 20 to 50 words.  \\

    Scene & Describe a \{scene\_label\} scene in the image.  
        Focus on the main elements typical of a \{scene\_label\}, such as objects, layout, colors, and lighting.  
        Avoid focusing on people or irrelevant details.  The description should be visual and vivid, 20 to 50 words. \\
    ImageNet-1k & Using the given label '\{label\}' as guidance, describe the image in as much detail as possible, focusing on objects, setting, colors, actions, and mood.  Incorporate elements related to the label.  Keep the description between 20 and 50 words. \\
    \bottomrule  
  \end{tabular}  
  }
\end{table}

\subsection{Captions Generation}
\label{3.2}
\zp{
To ensure the semantic fidelity of the generated captions to the original images, we utilize tailored templates for each semantic category, which then serve as prompts to guide QwenVL-chat~\cite{Qwen-VL} in its caption generation process.
} The caption templates are listed in Table~\ref{table-prompts}.
We design templates for all datasets to ensure that captioning  focuses on the main content of each image, except for COCO, where we directly use the provided captions. To ensure consistent generation quality and to match the number of generated and real images, we select 1,000 images for each semantic class, generate 1,000 captions, and instruct each generation model to produce two images per caption.
\gc{Representative image-caption pairs are provided in the supplemental material.}


\subsection{Image Generation}
\label{3.3}

\paragraph{Generation Models}
Text-to-image generation is increasingly becoming a primary method for image creation. 
Therefore, we carefully selected 31 representative commercial and open-source text-to-image models widely used in real-world scenarios. 
The distribution of these recent generative models is shown in Figure~\ref{fig:head}.

\gc{
For open-source models, we select those with relatively low video random access memory (VRAM) requirements to ensure ease of deployment on consumer-grade devices. These models span three major categories: LDM,  DiT, and AR.
}
\gc{
First, LDMs include the Stable Diffusion (SD) series and other variants based on LDM paradigm. We select the following models from the SD series: SD V1.5 \cite{Rombach_2022_CVPR}, SD V2.1 \cite{Rombach_2022_CVPR}, SDXL-base-V1.0 \cite{podell2023sdxl}, SD V3-medium~\cite{esser2024scaling}, and SD V3.5-medium \cite{esser2024scaling}. While SD V3 and V3.5 adopt DiT backbone, they are still widely recognized as part of the SD family. Additionally, we include other LDM-based models such as Playground V2.5 \cite{li2024playground}, and Kandinsky-3 \cite{vladimir-etal-2024-kandinsky}.
}
\gc{
Second, DiTs combine diffusion-based generation with transformer architectures to better capture global image context, leading to high-resolution and realistic outputs. In addition to SD V3, which follows the DiT design, we also include seven representative independently developed DiT-based models: FLUX.1-schnell \cite{flux2024}, PixArt-XL-2-1024-MS \cite{chen2023pixart}, CogView3-Plus \cite{zheng2024cogview3}, Hunyuan-DiT \cite{li2024hunyuandit}, HiDream-I1-fast \cite{hidream-l1}, FLUX.2-klein~\cite{flux2-klein}, and Z-Image~\cite{team2025zimage}. For simplicity, we roughly classify FLUX.2-klein as a DiT, although it is more precisely a flow-based model built around a flow transformer.
}
\gc{
Finally, AR models, which synthesize images in a token-by-token manner, have regained attention in the field due to their promising capabilities. We include three recent AR-based models: the visual text-to-image model Infinity-2B \cite{Infinity}, the multimodal model Janus-Pro \cite{chen2025janus}, and GLM-Image~\cite{glm-image}, which adopts a hybrid architecture with an AR generator and a DiT decoder.
}

For the commercial models, we focus on representative models with state-of-the-art performance and broad industry adoption. 
These can be divided into two categories: pure text-to-image generation models, including DALL·E 3~\cite{betker2023improving}, Midjourney~\cite{midjourney2022} (V5.2 and V6.0), Kling-image-V1~\cite{kling}, Ideogram-generate-V-1-TURBO~\cite{ideogram}, and Doubao-seedream-3.0-t2i~\cite{gao2025seedream}; and models integrated into widely deployed large multimodal frameworks, such as GPT-4o~\cite{achiam2023gpt}, GPT-Image-1~\cite{gpt-image-1}, GPT-Image-1.5~\cite{gpt-image-1_5}, Gemini-2.0-Flash~\cite{gemini-2-flash}, Gemini-2.5-Flash-Image~\cite{gemini-2_5-flash-image}, Gemini-3-Pro-Image~\cite{gemini-3-pro-image}, Doubao-seedream-5.0-Lite~\cite{seedream5_0_lite}, and Grok3-image~\cite{grok3}.

\paragraph{Image Generation}
Finally, we use captions generated in Section~\ref{3.2} to guide these generative models for image generation. 
To enhance realism and ensure styles closely match those of real images,  we apply negative prompts including terms such as ``cartoon, anime, drawing, painting, unrealistic, deformed features, blurry, low resolution, oversaturated, illustration, flat design, vector art, 2D, simple, graphic design.'' 
Implementation Details for generation models are as follows:
for models available on Hugging Face~\cite{huggingface},  we deploy them and generate images using the Python Diffusers pipeline~\cite{von-platen-etal-2022-diffusers}. SD V1.5 and FLUX.1-schnell produce images at the resolution of 512×512, SD V2.1 and  Kandinsky-3 at 768×768 and the rest at 1024×1024. 
Other open-source models are deployed via their GitHub repositories. Janus-pro generates images at 384×384, Hunyuan-DiT , Infinity-2B and HiDream-I1-fast at 1024×1024. 
Commercial models generate images through API calls. DALL·E 3, GPT-4o, GPT-Image-1, GPT-Image-1.5, Gemini-2.5-Flash-Image, Gemini-3-Pro-Image, Doubao-seedream-3.0-t2i, Midjourney V5.2, Midjourney V6.0, Kling-image-V1, and Ideogram-generate-V-1-TURBO all produce images at a resolution of 1024×1024. Doubao-seedream-5.0-Lite generates images at 1920×1920. Grok3-image generates images at 1024×768, and Gemini-2.0-Flash generates images at varying resolutions, with the longer side fixed at 1024.

Sample generated images are shown in Figure~\ref{fig:sample}, where the vertical axis indicates the semantic classes, and the horizontal axis corresponds to a subset of generative models used in our dataset construction. Additional visualizations are provided in Section~\ref{Additional visualization} of the supplemental material.

\section{Experimental Setup}
\paragraph{Attribution Methods}
As we mention in Section~\ref{related work}, existing methods can be categorized into four classes. For each class, we select a few representative methods.
For the spatial and frequency feature-based methods, we select ResNet-50~\cite{he2016deep} pretrained on ImageNet~\cite{ILSVRC15}; DCT-CNN~\cite{frank2020leveraging}, which exploits the frequency domain; the method from~\cite{cioni2024clip}, which utilizes a CLIP model for feature extraction and a logistic regressor for classification; and HiFi-Net~\cite{guo2023hierarchical} (without the localization module, since it is not applicable to this task). We also reproduce method from~\cite{li2024handcrafted}, which employs a handcrafted filter combined with the CLIP feature extractor. To obtain better and more stable results, we replace the reference set average distance classifier with a neural network classifier.
For the data augmentation-based methods, we choose DNA-Det~\cite{yang2022deepfake} with two phases of training and RepMix~\cite{bui2022repmix}.
For the image patch-based methods, we leverage PatchForensics~\cite{chai2020makes},  SSP~\cite{chen2024single}, and reproduce PatchCraft~\cite{zhong2023patchcraft}.
For the disentangled representation learning-based methods, we reproduce GFD~\cite{yang2021learning} and adapt UCF~\cite{yan2023ucf} from a detection task to an attribution task. We also incorporate two SOTA few-shot detectors, FSD~\cite{wu2025fewshotlearnergeneralizesaigenerated} and Omni-DFA~\cite{wu2025omnidfaunifiedframeworkopen}, and tailor them to the attribution task.

\paragraph{Evaluation Settings}
\label{3.4}
\zp{
We evaluate these methods in two tasks: a standard balanced-split task and a semantic-split task. 
}
For the standard balanced-split task, we typically divide the dataset into three subsets: training, validation and testing, with a ratio of 8:1:1. We balance the labels to ensure an equal distribution across all classes. For the semantic-split task, the train-validation and test sets are split based on semantic classes. Specifically, the training and validation sets contain samples from only one semantic class, split in a 9:1 ratio, while the test set consists of samples from the remaining classes. This kind of split significantly reduces the utilization of semantic information and challenges the capability to extract semantic-irrelevant fingerprints of generative models. For this task, we design three sub-tasks that use cat faces from AFHQ, human faces from FFHQ, and bedrooms from LSUN for training and validation, respectively. 
In real-world scenarios, images often undergo various types of degradation, such as compression, cropping, and blurring during transmission across social media platforms. 
Therefore, to simulate these conditions and evaluate the robustness of each attributor, we apply six degradation levels that include various types—downsampling, JPEG compression, and Gaussian blur—with different intensities, following the protocol in GenImage~\cite{zhu2023genimage}.
Regarding metrics, we use accuracy to measure the performance. During evaluation, we resize and center-crop images to 256×256 and train each attributor equally for 10 epochs on the standard task and 15 epochs on the semantic task, ensuring that all can fit on the training set. We then select the best checkpoint based on validation set performance and evaluate on the test set.

\begin{table*}[t]
  \centering
  \caption{Comparison of different methods on both the standard balanced-split task and the semantic-split task. For the standard task, DS denotes downsampling, JPEG denotes JPEG compression, and Blur represents Gaussian blur. For the semantic task, each cell reports clean accuracy (left) and the average degraded accuracy across six degradation levels (right).}
  \resizebox{\linewidth}{!}{
    \begin{tabular}{l|cccccccc|cccc}
    \toprule
    \multicolumn{1}{l|}{\multirow{4}[6]{*}{Method}} & \multicolumn{8}{c|}{Standard Task}                             & \multicolumn{4}{c}{Semantic Task} \\
\cmidrule{2-13}          & \multicolumn{1}{c}{\multirow{3}[4]{*}{\makecell[c]{Clean\\Acc. (\%)}}} & \multicolumn{7}{c|}{Degraded Acc.}                    & \multicolumn{1}{c}{\multirow{3}[4]{*}{\makecell[c]{Training on \\ Cat \\ Acc.}}} & \multicolumn{1}{c}{\multirow{3}[4]{*}{\makecell[c]{Training on \\ Human Face \\ Acc.}}} & \multicolumn{1}{c}{\multirow{3}[4]{*}{\makecell[c]{Training on\\ Bedroom \\ Acc.}}} & \multicolumn{1}{c}{\multirow{3}[4]{*}{\makecell[c]{Avg \\ Acc. }}} \\
\cmidrule{3-9}          &       & \multicolumn{1}{l}{DS} & \multicolumn{1}{l}{DS} & \multicolumn{1}{l}{JPEG} & \multicolumn{1}{l}{JPEG} & \multicolumn{1}{l}{Blur} & \multicolumn{1}{l}{Blur} & \multicolumn{1}{l|}{Avg} &       &       &       &  \\
          &       & \multicolumn{1}{l}{(0.5x)} & \multicolumn{1}{l}{(0.25x)} & \multicolumn{1}{l}{(q=65)} & \multicolumn{1}{l}{(q=30)} & \multicolumn{1}{l}{($\sigma = 3$)} & \multicolumn{1}{l}{($\sigma = 5$)} & \multicolumn{1}{l|}{Acc.} &       &       &       &  \\
    \midrule
    ResNet-50~\cite{he2016deep} & 89.7 & 89.2 & \textbf{83.6} & 77.3 & 60.3 & 58.8 & \underline{36.4} & \underline{67.6} & 21.9/19.1 & 26.3/23.5 & 18.0/14.6 & 22.1/19.1 \\
    DCT-CNN~\cite{frank2020leveraging} & 75.2 & 66.2 & 45.0 & 27.7 & 20.1 & 6.0 & 1.2 & 27.7 & 23.2/13.7 & 34.0/15.9 & 30.9/15.2 & 29.4/14.9 \\
    HiFi-Net~\cite{guo2023hierarchical} & 96.8 & 88.9 & 60.1 & 16.0 & 13.5 & 36.7 & 10.2 & 37.6 & 50.8/24.8 & 61.9/24.0 & 56.4/25.6 & 56.4/24.8 \\
    \citet{li2024handcrafted} & 96.6 & 89.5 & 65.0 & 40.0 & 36.0 & 47.1 & 11.8 & 48.2 & 44.9/19.3 & 51.1/20.2 & 43.2/17.6 & 46.4/19.0 \\
    \midrule
    DNA-Det~\cite{yang2022deepfake} & 92.9 & 87.2 & 56.4 & 14.8 & 12.9 & 44.7 & 12.1 & 38.0 & \underline{68.6}/28.5 & \textbf{74.6}/26.4 & \underline{70.6}/27.1 & \textbf{71.3}/27.3 \\
    RepMix~\cite{bui2022repmix} & 90.0 & 89.2 & \underline{80.2} & \underline{89.8} & \underline{81.0} & 37.3 & 19.0 & 66.1 & 66.3/\textbf{50.3} & 59.9/\textbf{49.1} & 66.6/\textbf{48.8} & 64.3/\textbf{49.4} \\
    \midrule
    PatchForensics~\cite{chai2020makes} & 93.8 & 89.2 & 60.2 & 14.6 & 12.2 & 44.2 & 10.2 & 38.4 & \textbf{69.1}/\underline{32.6} & \underline{67.2}/29.7 & \textbf{73.4}/\underline{35.1} & \underline{69.9}/\underline{32.5} \\
    PatchCraft~\cite{zhong2023patchcraft} & 73.5 & 5.3 & 3.9 & 32.9 & 13.1 & 8.7 & 7.2 & 11.9 & 44.5/8.4 & 39.0/8.1 & 39.1/7.1 & 40.9/7.9 \\
    SSP~\cite{chen2024single} & 89.0 & 76.5 & 39.3 & 12.7 & 7.7 & 43.4 & 15.1 & 32.5 & 61.2/24.9 & 62.0/23.5 & 61.3/24.2 & 61.5/24.2 \\
    \midrule
    GFD~\cite{yang2021learning} & 87.7 & 79.8 & 53.5 & 70.9 & 44.7 & 13.4 & 4.7 & 44.5 & 45.9/\underline{34.0} & 25.7/22.3 & 43.0/25.7 & 37.2/27.3 \\
    UCF~\cite{yan2023ucf} & 95.9 & 94.0 & 75.2 & 40.4 & 24.3 & 43.3 & 9.0 & 47.7 & 16.9/13.5 & 21.8/17.6 & 17.9/12.4 & 18.9/14.5 \\
    \midrule
    FSD~\cite{wu2025fewshotlearnergeneralizesaigenerated} & \textbf{97.5} & \underline{95.5} & 71.5 & 62.4 & 62.4 & \underline{68.1} & 26.6 & 64.4 & 35.3/23.6 & 36.8/21.4 & 34.3/20.3 & 35.5/21.7 \\
    Omni-DFA~\cite{wu2025omnidfaunifiedframeworkopen} & \underline{97.2} & \textbf{96.4} & 79.0 & \textbf{96.1} & \textbf{83.8} & \textbf{95.6} & \textbf{89.0} & \textbf{90.0} & 36.5/29.1 & 43.4/\underline{35.7} & 39.5/29.6 & 39.8/31.5 \\
    \midrule
    Avg & 90.4 & 80.5 & 59.5 & 45.8 & 36.3 & 42.1 & 19.4 & 47.3 & 45.0/24.7 & 45.8/24.5 & 46.1/23.2 & 45.6/24.2 \\
    \bottomrule
    \end{tabular}}
  \label{tab:main}
\end{table*}

\section{Results}
In this section, we discuss the performances of SOTA attributors on ImageAttributionBench. 
We review the results on standard split task with both clean and degraded testing data. Furthermore, we evaluate performance on semantic-split task, and analyze the correlation between accuracy and semantic  similarity of the training and testing subsets.
\paragraph{Performance on Standard Balanced-Split Task}

Following evaluation settings in Section~\ref{3.4}, we first conduct experiments on the standard task, and results are shown in the left half of Table~\ref{tab:main}. 
On clean images, most attributors exceed 80\%, with an average accuracy of 90.4\%.
Notably, FSD, originally proposed for few-shot fake image detection, attains the highest performance at 97.5\%, while even the pretrained ResNet-50 reaches 89.7\%.
This indicates that the availability of full semantic information in the standard task simplifies the attribution process.

However, under the more practical scenario of image degradation, the average accuracy of all attributors drops sharply from 90.4\% to 47.3\%.
This indicates that the degradation of spatial details, frequency structures, and edge information poses a considerable challenge for current attributors.
In contrast, superior methods like Omni-DFA and RepMix use data augmentation in data processing or leverage feature-mixing strategies, leading to better performance under such conditions.
This underscores the importance of developing more robust feature extraction techniques to handle image degradation effectively.


\begin{figure}[t]
  \centering
  \begin{subfigure}[t]{0.48\linewidth}
    \centering
    \includegraphics[width=\linewidth]{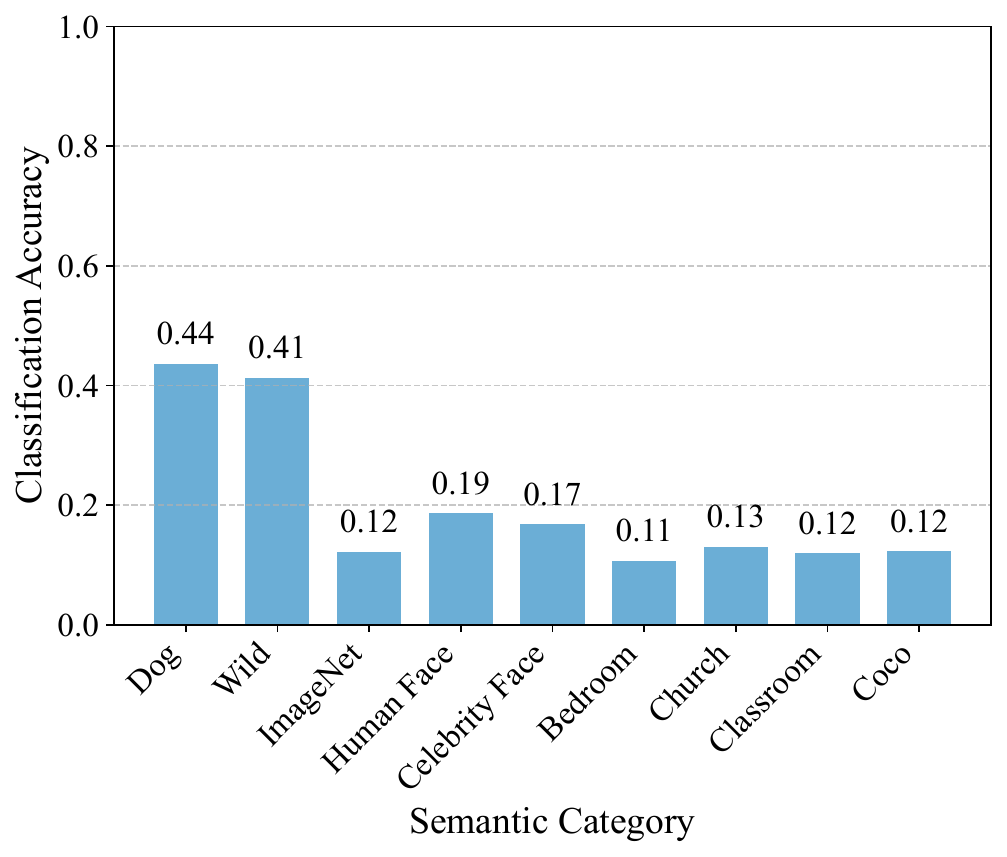}
    \caption{Accuracy results of ResNet-50 trained on ``Cat.''}
    \label{fig:resnet50}
  \end{subfigure}
  \begin{subfigure}[t]{0.48\linewidth}
    \centering
    \includegraphics[width=\linewidth]{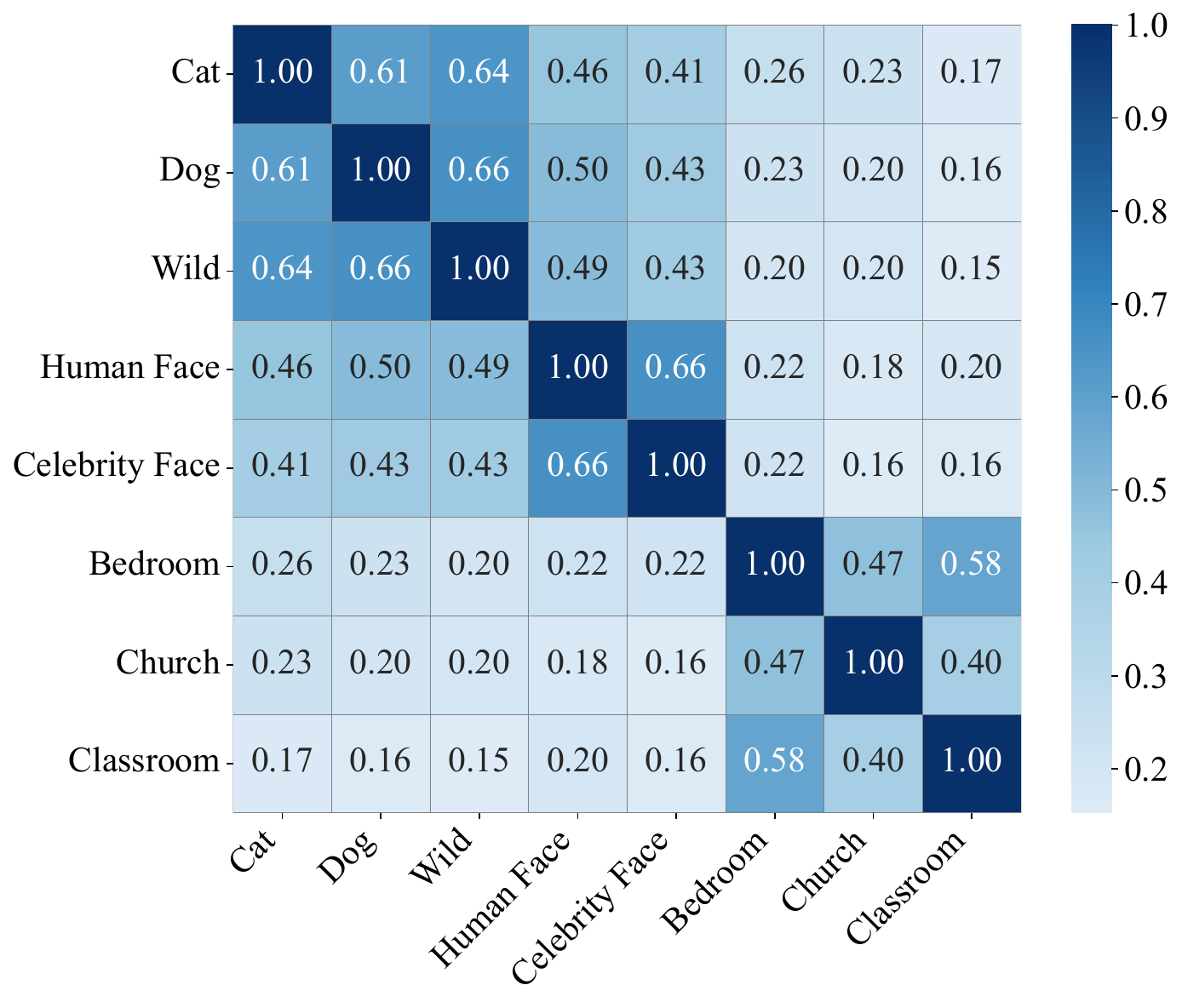}
    \caption{Semantic similarity matrix.}
    \label{fig:similarity_matrix}
  \end{subfigure}
  \caption{Performance of ResNet-50 and semantic similarity matrix. (a) Accuracy results for each semantic category obtained from ResNet-50 trained on ``Cat.'' (b) Semantic similarity matrix across different subsets.}
  \label{fig:resnet50_similarity}
\end{figure}

\paragraph{Performance on Semantic-Split Task}
\zp{
In the semantic task, the dataset is divided into 10 semantically distinct subsets (e.g., ``Cat'', ``Human face'', ``Bedroom''). We then perform training on one subset at a time and evaluate these methods on the remaining nine. The results of this evaluation are presented in the right half of Table~\ref{tab:main}.
}
\zp{Compared to the standard task, the semantic task results in a significant drop in clean average performance, from 90.4\% to around 45.6\%. 
Notably, the best-performing methods on the standard task, FSD and Omni-DFA, yield poor performance on the semantic task, with accuracies below 40\%, exhibiting significant performance drops of over 50 percentage points.
This suggests that the separation of semantic categories can have a similarly detrimental effect on performance as image degradation.
DNA-Det and RepMix show comparatively strong performance on the semantic task.
This highlights that data augmentations, such as feature mixing and pretraining with image transformations, can improve generalization ability even when the model is trained on images from a single semantic category. 
Despite this robustness, DNA-Det and RepMix still experience clear performance declines of 21.6 and 25.7 percentage points, respectively, from the standard clean task to the semantic clean task.}
PatchForensics and SSP achieve moderately high accuracy, demonstrating the potential of local patch information in cross-semantic attribution. 
In contrast, PatchCraft does not achieve similar results, as its strategy is specifically designed for detection rather than attribution.
\zp{Under image degradation, the semantic task shows a further 21.4\% drop in average performance, from 45.6\% to 24.2\%.
This highlights that existing attribution methods still have significant room for improvement in more challenging yet practical scenarios.}

\begin{figure*}[t]
  \centering
\includegraphics[width=\textwidth]{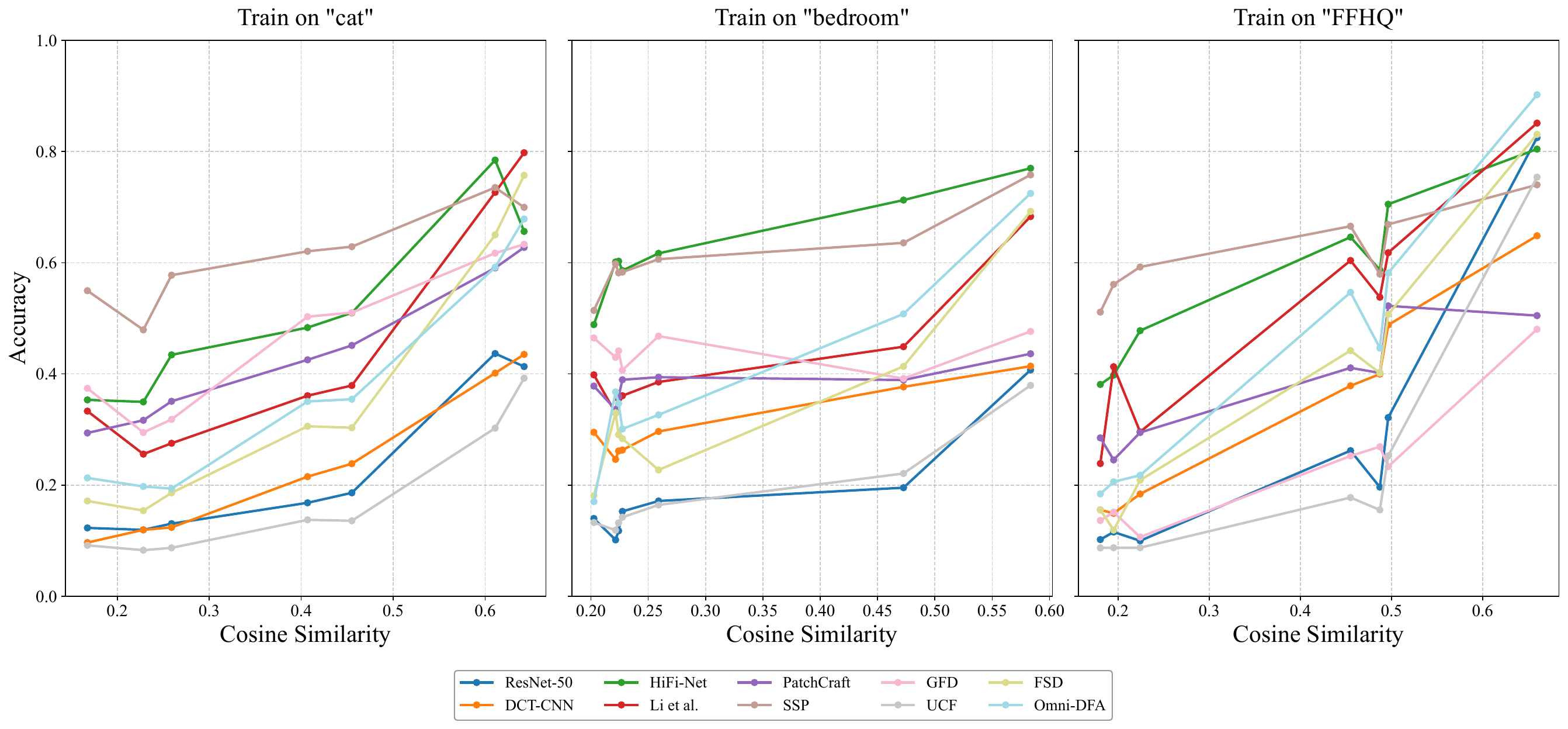}
  \caption{Accuracy against semantic similarity for different attribution models across various training settings, i.e., training on ``cat'', ``bedroom'', and ``FFHQ''.}
  \label{fig:three}
\end{figure*}

\paragraph{Correlation Between Accuracy and Semantic Similarity}   
\zp{To investigate the relationship between attribution accuracy and semantic similarity, we report per-subset accuracy in Figure~\ref{fig:resnet50} using ResNet-50 trained on the ``cat'' category, and show the category similarity matrix in Figure~\ref{fig:similarity_matrix}, where 100 randomly selected captions are encoded into sentence embeddings and their cluster centers compared using cosine similarity.
From these results, we observe that attribution accuracy declines as the semantic similarity to the training category ``cat'' decreases. 
For instance, ``dog'' and ``wild'' categories, which are semantically close to ``cat'', achieve relatively high accuracy.
\zp{To further illustrate this relationship more explicitly, we plot accuracy against similarity across various training settings in Figure~\ref{fig:three}.}
As shown, there is a clear positive correlation between accuracy and semantic similarity.
This suggests that existing methods benefit from semantic correlations between training and testing splits in current datasets.
Therefore, semantic separation in ImageAttributionBench is crucial for a more rigorous and realistic evaluation of attribution performance.}

\section{Limitations}
\label{limitation}
\zp{
Our proposed ImageAttributionBench introduces a novel resource encompassing state-of-the-art generative model architectures and presents challenges to attribution tasks. 
While acknowledging that our initial release is not exhaustive, we have focused on establishing a strong foundation. 
Specifically, our semantic coverage includes several commonly studied categories in attribution, providing a targeted benchmark for current research. Similarly, we have adopted a photorealistic style to align with real-world imagery. Recognizing the resource constraints of initial data generation, we provide 2,000 high-quality images per model per semantic category, offering a substantial dataset for practical evaluation. 
We envision significant potential for future extensions, including incorporation of a broader spectrum of semantic categories (e.g., drawing inspiration from the diversity of ImageNet), exploration of artistic and other stylistic variations, and scaling to larger dataset sizes to facilitate more comprehensive analyses of attribution methods.
}

\section{Conclusion}
This paper introduces ImageAttributionBench, a benchmark specialized for AI-synthesized image attribution. 
ImageAttributionBench surpasses previous datasets in both the number and advancement of generative model architectures it includes, while maintaining semantic diversity and balance across models. 
We conduct two main experiments on ImageAttributionBench, namely the standard balanced-split task and the semantic-split task, and the results demonstrate the limitations of existing attribution methods in terms of robustness to degradation and semantic generalization. 
Additionally, we provide a quantitative analysis of the correlation between accuracy and semantic similarity, showing the need for improved attribution methods to overcome semantic entanglement. 
We hope that ImageAttributionBench will inspire the AI community to advance attribution techniques that are robust, semantically generalizable, and applicable in real-world settings.

\small{
\bibliography{ia}
}

\appendix

\clearpage
\setcounter{page}{1}

\section{Ethical Data Curation and Protection}
\label{ethic}
\label{sec:ethic}
Given the inclusion of human faces in our dataset, we take the ethical considerations surrounding face-related images seriously. To address these issues, we apply demographic balancing during data sampling and perform Personally Identifiable Information (PII) filtering in caption generation.

\begin{figure}[h]
  \centering
  \begin{subfigure}[t]{0.44\linewidth}
    \centering
    \includegraphics[width=\linewidth]{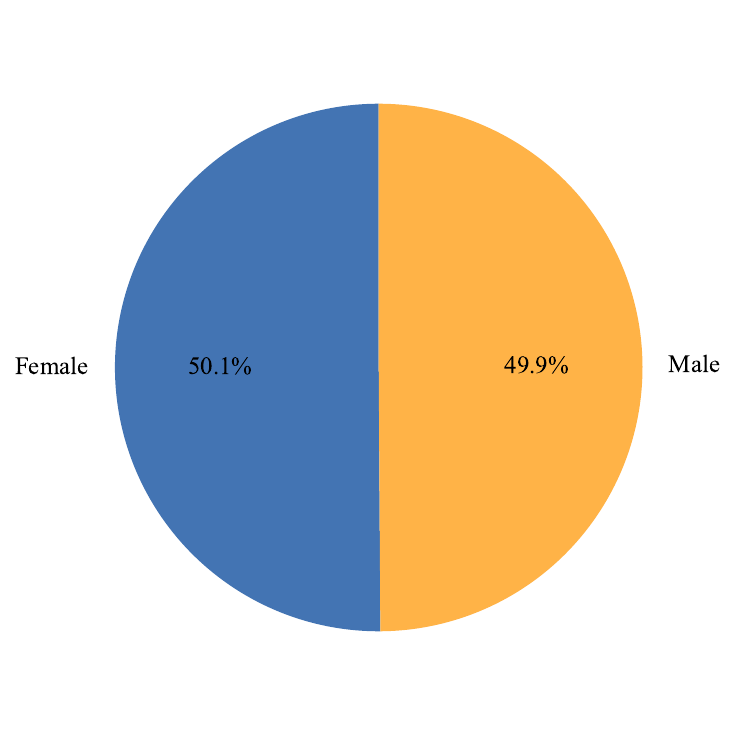}
    \caption{Gender distribution of the face-related subset.}
    \label{fig:pie_gender}
  \end{subfigure}
  \vspace{4pt}
  \begin{subfigure}[t]{0.54\linewidth}
    \centering
    \includegraphics[width=\linewidth]{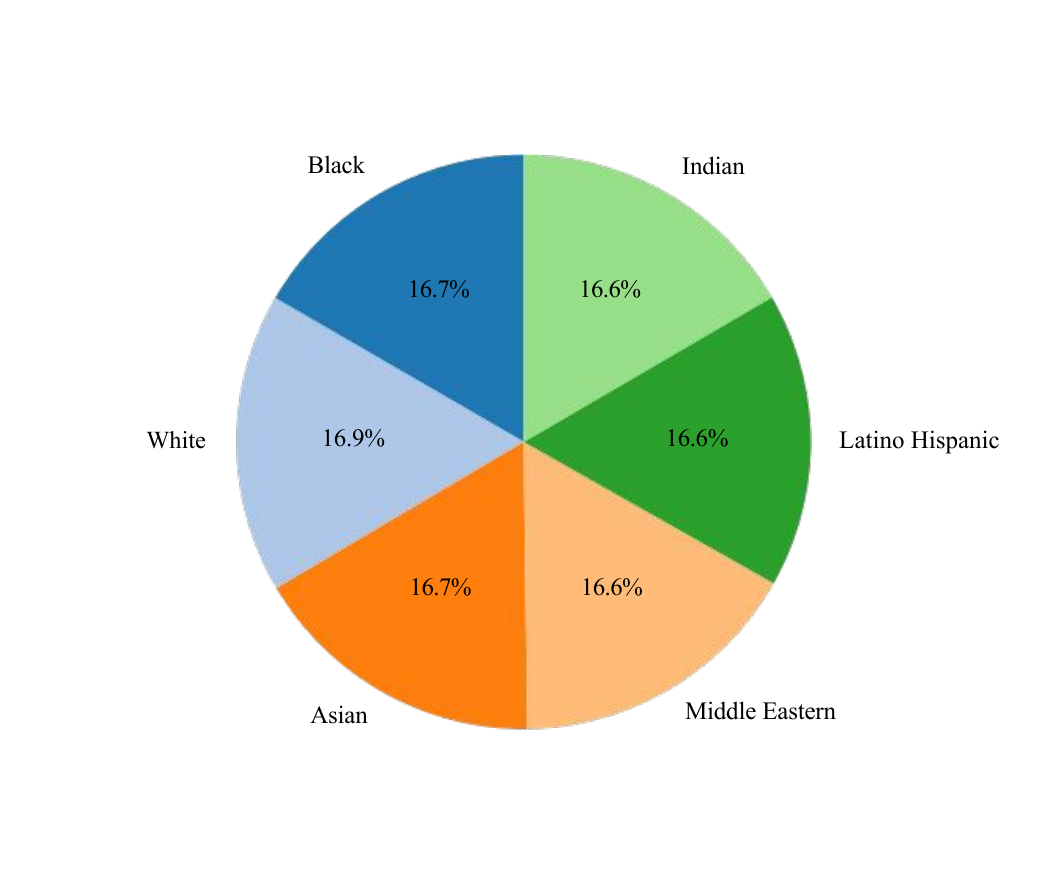}
    \caption{Racial distribution of the face-related subset.}
    \label{fig:pie_race}
  \end{subfigure}
  \caption{Demographic composition of the real face-related subset. 
  (a) Gender distribution showing the ratio of male and female faces. 
  (b) Racial distribution across six demographic categories. 
}
  \label{fig:demographic_distribution}
\end{figure}

For demographic balancing, we make every effort to ensure fair gender and race representation within the dataset. To achieve this, we employ FairFace~\cite{karkkainenfairface}, a gender and race detector, to sample a representative subset of real images (e.g., FFHQ and CelebA-HQ) with balanced gender and racial distributions. The resulting demographic distribution is illustrated in Figure~\ref{fig:demographic_distribution}. 

For PII filtering, we design privacy-safe prompts for human face-related caption generation, emphasizing that no private information should appear in the captions produced by the VLM captioner. After caption generation, we employ the large language model Llama-3.1-8B~\cite{patterson2022carbon} to further detect and remove privacy-violating content in the captions. This procedure ensures that all captions, and consequently the generated images, remain privacy-safe. The detailed prompt templates are shown in Table~\ref{table-prompts}.

\section{Choice of Vision-Language Model (VLM)}
In our dataset construction pipeline, we adopt QwenVL-Chat as the captioning model, as it represented a state-of-the-art VLM at the time of our experiments and consistently produced high-quality prompts that satisfied our dataset requirements. To validate this choice, we conducted comparative experiments using several vision-language models (VLMs), as presented in Table~\ref{tab:vlm_similarity}. The results show that both text and image similarities between the generated prompts and the corresponding images remain consistently high across models, suggesting that the specific choice of VLM has a relatively limited influence on prompt quality. This observation can be attributed to the large-scale and diverse training data used by modern VLMs, which endow them with strong generalization capabilities.

\begin{table}[t]
  \centering
  \caption{Text and image similarity of different VLMs with QwenVL-Chat. 
  Text similarity is measured using Sentence-BERT similarity, 
  while image similarity is measured using CLIP similarity.}
  \label{tab:vlm_similarity}
  \resizebox{\linewidth}{!}{
    \begin{tabular}{lcc}
      \toprule
      Model & Text Similarity & Image Similarity \\
      \midrule
      Qwen2.5-VL-7B~\cite{qwen2.5-VL} & 0.7521 & 0.9704 \\
      Llama-3.2-11B-Vision-Instruct~\cite{patterson2022carbon} & 0.7390 & 0.9813 \\
      InternVL3-8B~\cite{wang2025internvl3_5} & 0.7388 & 0.9756 \\
      \bottomrule
    \end{tabular}
  }
\end{table}

\section{Image Generation Details}
The dataset construction is conducted on an RTX 3090 GPU with 24 GB of memory. The generation parameters of the open-source models are listed in Table~\ref{tab:open_source_model_params}.

\begin{table}[t]
\centering
\caption{Generation parameters of open source models.}
\label{tab:open_source_model_params}
\resizebox{\linewidth}{!}{ 
\begin{tabular}{lcccc}
\toprule
Model Name               & num\_inference\_steps & guidance\_scale & width/height & enable\_model\_cpu\_offload \\ 
\midrule
SD V1.5~\cite{Rombach_2022_CVPR}                 & 30    & 7.5              & 512          & No       \\ 
SD V2.1~\cite{Rombach_2022_CVPR} & 30    & 7.5              & 768          & No       \\ 
SD V3-medium~\cite{esser2024scaling}            & 28    & 7.0              & 1024         & No       \\ 
SD V3.5-medium~\cite{esser2024scaling}          & 40    & 4.5              & 1024         & No       \\ 
SDXL-base-V1.0~\cite{podell2023sdxl}         & 30    & 5.0              & 1024         & No       \\ 
FLUX.1-schnell~\cite{flux2024}         & 4     & 3.5              & 512          & Yes      \\ 
Kandinsky-3~\cite{vladimir-etal-2024-kandinsky}            & 25    & 4.0              & 768          & Yes      \\ 
PixArt-XL-2-1024-MS~\cite{chen2023pixart}      & 20    & 6.0              & 1024         & No       \\ 
Playground V2.5~\cite{li2024playground}         & 50    & 3.0              & 1024         & Yes      \\ 
CogView3-Plus~\cite{zheng2024cogview3}           & 50    & 7.0              & 1024         & Yes      \\ 
HiDream-I1-fast~\cite{hidream-l1}         & 16    & 0.0     & 1024  & Yes \\
Hunyuan-DiT~\cite{li2024hunyuandit}         & 35    & 6.0     & 1024  & Yes \\
Z-Image ~\cite{team2025zimage}         & 9    & 0.0     & 1024  & Yes \\
FLUX.2-klein ~\cite{flux2-klein}         & 4    & 1.0     & 1024  & Yes \\
GLM-Image ~\cite{glm-image}         & 50    & 1.5     & 1024  & Yes \\
\bottomrule
\end{tabular}
}
\end{table}

\section{Training Details}
\paragraph{Training Subset Selection for Semantic Task} For the semantic-split task, we select three subsets: ``cat,'' ``bedroom'' and ``human face'' (from FFHQ) as representative subsets for training, based on the following reasons: (1) All semantic classes can be categorized into three broad groups: scenes (including ``church,'' ``bedroom'' and ``classroom''), animals (including ``dog,'' ``cat'' and ``wild''), and human faces (including ``celebrity face'' from CelebA-HQ and ``human face'' from FFHQ). For each group, we select one subset as its representative. (2) For each group, we choose the most commonly used semantic concept: ``bedroom'' for scenes, ``cat'' for animals, and FFHQ ``human face'' for human faces, as FFHQ offers greater age diversity compared to CelebA-HQ.

\paragraph{Training Settings of Attributors}

\begin{table}[t]
  \centering
  \caption{Multi level label mapping for HiFi-Net.}
  \label{tab:model_level_mapping_transposed}
  \resizebox{\linewidth}{!}{
  \begin{tabular}{ccccc}
  \toprule
  Model Name            & Level 0 & Level 1 & Level 2 & Level 3 \\
  \midrule
  GPT4-o                & 0 & 0 & 0 & 0 \\
  DALL·E 3              & 0 & 0 & 0 & 11 \\
  Gemini-2.0-flash-exp-image-generation & 0 & 0 & 0 & 12 \\
  Grok3-image           & 0 & 0 & 0 & 13 \\
  Ideogram-generate-V-1-TURBO & 0 & 0 & 0 & 16 \\
  Kling-image-V1        & 0 & 0 & 0 & 19 \\
  Midjourney V5.2       & 0 & 0 & 0 & 20 \\
  Midjourney V6.0       & 0 & 0 & 0 & 21 \\
  Doubao-Seedream-3.0-t2i & 0 & 0 & 0 & 26 \\
  Doubao-Seedream-5.0-lite & 0 & 0 & 0 & 27 \\
  Gemini-2.5-flash-image & 0 & 0 & 0 & 28 \\
  Gemini-3-pro-image    & 0 & 0 & 0 & 29 \\
  GPT-image-1           & 0 & 0 & 0 & 30 \\
  GPT-image-1.5         & 0 & 0 & 0 & 31 \\
  SD V1.5               & 0 & 1 & 1 & 6 \\
  SD V2.1               & 0 & 1 & 1 & 7 \\
  SD V3-medium          & 0 & 1 & 1 & 8 \\
  SD V3.5-medium        & 0 & 1 & 1 & 9 \\
  SDXL-base-V1.0        & 0 & 1 & 1 & 10 \\
  FLUX.1-schnell        & 0 & 1 & 2 & 2 \\
  Kandinsky-3           & 0 & 1 & 2 & 3 \\
  PixArt-XL-2-1024-MS   & 0 & 1 & 2 & 4 \\
  Playground V2.5       & 0 & 1 & 2 & 5 \\
  FLUX.2-Klein          & 0 & 1 & 2 & 23 \\
  Z-Image               & 0 & 1 & 2 & 25 \\
  CogView3-Plus         & 0 & 1 & 3 & 1 \\
  HiDream-I1-fast       & 0 & 1 & 3 & 14 \\
  Hunyuan-DiT           & 0 & 1 & 3 & 15 \\
  Infinity-2B           & 0 & 1 & 4 & 17 \\
  Janus-Pro             & 0 & 1 & 4 & 18 \\
  GLM-Image             & 0 & 1 & 4 & 24 \\
  real                  & 1 & 2 & 5 & 22 \\
  \bottomrule
  \end{tabular}
  }
  \end{table}

We utilize different training settings for different attributors, as shown below. For any settings not mentioned, the default configuration is applied: training for 10 epochs for the standard task and 15 epochs for the semantic  task, using the Adam optimizer with an initial learning rate of 0.001, a batch size of 32, and cross-entropy loss by default.

\textbf{ResNet50}~\cite{he2016deep} We use a pretrained model from the Python package torchvision.models as backbone, replacing the classifier with identity and adding a linear layer for classification. The number of training epochs is 20 for standard task.

\textbf{DCT-CNN}~\footnote{https://github.com/RUB-SysSec/GANDCTAnalysis/}~\cite{frank2020leveraging} We apply the DCT transform to images following the official repository, and use a simple CNN with four convolutional layers as the backbone for attribution.

\textbf{Cioni et al.}~\cite{cioni2024clip} We use pretrained CLIP Model~\footnote{https://huggingface.co/laion/CLIP-ViT-H-14-laion2B-s32B-b79K} as the backbone. With the backbone frozen, we train a logistic regressor using L2 loss and the LBFGS solver~\cite{liu1989limited}, with a maximum of 1000 training iterations.

\textbf{HiFi-Net}~\footnote{https://github.com/CHELSEA234/HiFi\_IFDL}~\cite{guo2023hierarchical} We utilize NLCDetection and HighResolutionNet from the official repository, designing the multi-level labels as shown in table~\ref{tab:model_level_mapping_transposed} based on the attributes of generative models (e.g., open source or commercial, DiT, AR, or LDM-based, etc.). The training batch size is set to 8. 

\textbf{Li et al.}~\cite{li2024handcrafted} We reproduce MHF and DEFL based on the paper. Additionally, we employ a neural network classifier and combine dual contrastive loss with cross-entropy loss for model fitting. The batch size is set to 8, and the number of epochs is 10 for both the standard and semantic tasks.

\textbf{DNA-Det}~\footnote{https://github.com/ICTMCG/DNA-Det}~\cite{yang2022deepfake} We adopt the main architecture from the official repository and train for 10 epochs in both the pre-training phase and the attribution training phase for each task.

\textbf{RepMix}~\footnote{https://github.com/tubui/image\_attribution}~\cite{bui2022repmix} We adopt the complete network architecture from the official repository. Notably, we forgo data augmentation in the data processing stage. 

\textbf{PatchForensics}~\footnote{https://github.com/chail/patch-forensics}~\cite{chai2020makes} We use Xception as the backbone, which is the default model in the official repository. For the accuracy metric, we select acc\_D\_voted (where the predicted label is the one with the highest vote count across all patches). The number of epochs is set to 20 for both tasks.

\textbf{PatchCraft}~\cite{zhong2023patchcraft} We strictly reproduce the method as described in the paper, with the only modification being the classifier head changed from single-class to multi-class classification.

\textbf{SSP}~\footnote{https://github.com/bcmi/SSP-AI-Generated-Image-Detection}~\cite{chen2024single} We adopt the data processing strategy from the repository to extract the simplest patches and utilize a pretrained ResNet-50 for attribution. Both tasks are trained for 10 epochs each.

\textbf{GFD}~\cite{yang2021learning} We reproduce this method following the paper, employing a U-Net encoder-decoder for fingerprint extraction, a PatchGAN~\footnote{https://github.com/He-jerry/PatchGAN/blob/master/network.py} discriminator and a pretrained ResNet-50 as auxiliary classifier. In each training step, we first train the U-Net fingerprint generator, then jointly train the discriminator and classifier. Batch size is set to 16.

\textbf{UCF}~\footnote{https://github.com/SCLBD/DeepfakeBench/}~\cite{yan2023ucf} We adopt the entire network architecture from official repository, modifying only the attribution head and swapping the loss weights between detection and attribution. The number of epochs is set to 10 for both tasks.

\textbf{FSD}~\footnote{https://github.com/teheperinko541/Few-Shot-AIGI-Detector}~\cite{wu2025fewshotlearnergeneralizesaigenerated}
  We reproduce this method following the few-shot attribution setting in the paper, but use the
  multiclass implementation in our codebase for unified comparison. Concretely, we adopt an
  ImageNet-pretrained ResNet-50 backbone, replace the final classification layer with a 1024-
  dimensional embedding head, and attach a linear attribution classifier for source prediction. We
  use Adam optimizer with learning rate \(1\times10^{-4}\) and a step scheduler with decay factor
  0.5 every 5 epochs. The batch size is set to 32, and both the standard balanced-split task and the
  semantic-split task are trained for 10 epochs each.

  \textbf{Omni-DFA}~\cite{wu2025omnidfaunifiedframeworkopen} We reproduce this method with a dual-branch ConvNeXt-Small
  architecture, where one branch processes global views and the other processes local views, and the
  fused features are projected to a 128-dimensional normalized embedding space for attribution.
  During training, we maintain learnable class prototypes and a real-image center, and optimize the
  model using supervised contrastive loss together with a center regularization term for real
  images. We use AdamW optimizer with learning rate \(2\times10^{-5}\), weight decay 0.01, and a
  cosine warmup schedule with 2 warmup epochs. The batch size is set to 144, and both the standard
  balanced-split task and the semantic-split task are trained for 20 epochs each.

\section{Extended Results and Visualizations}
In this section, we present additional result tables, confusion matrices and frequency spectra of generated images.

\paragraph{Results on Three Semantic Tasks}
Results with accuracy for each degradation level across different attribution methods training on three semantic tasks (i.e., ``cat,'' ``bedroom'' and ``human face'') are presented in Table~\ref{tab:sp1}, Table~\ref{tab:sp2} and Table~\ref{tab:sp3} respectively.

\begin{figure*}[h]
  \centering
  \includegraphics[width=\textwidth]{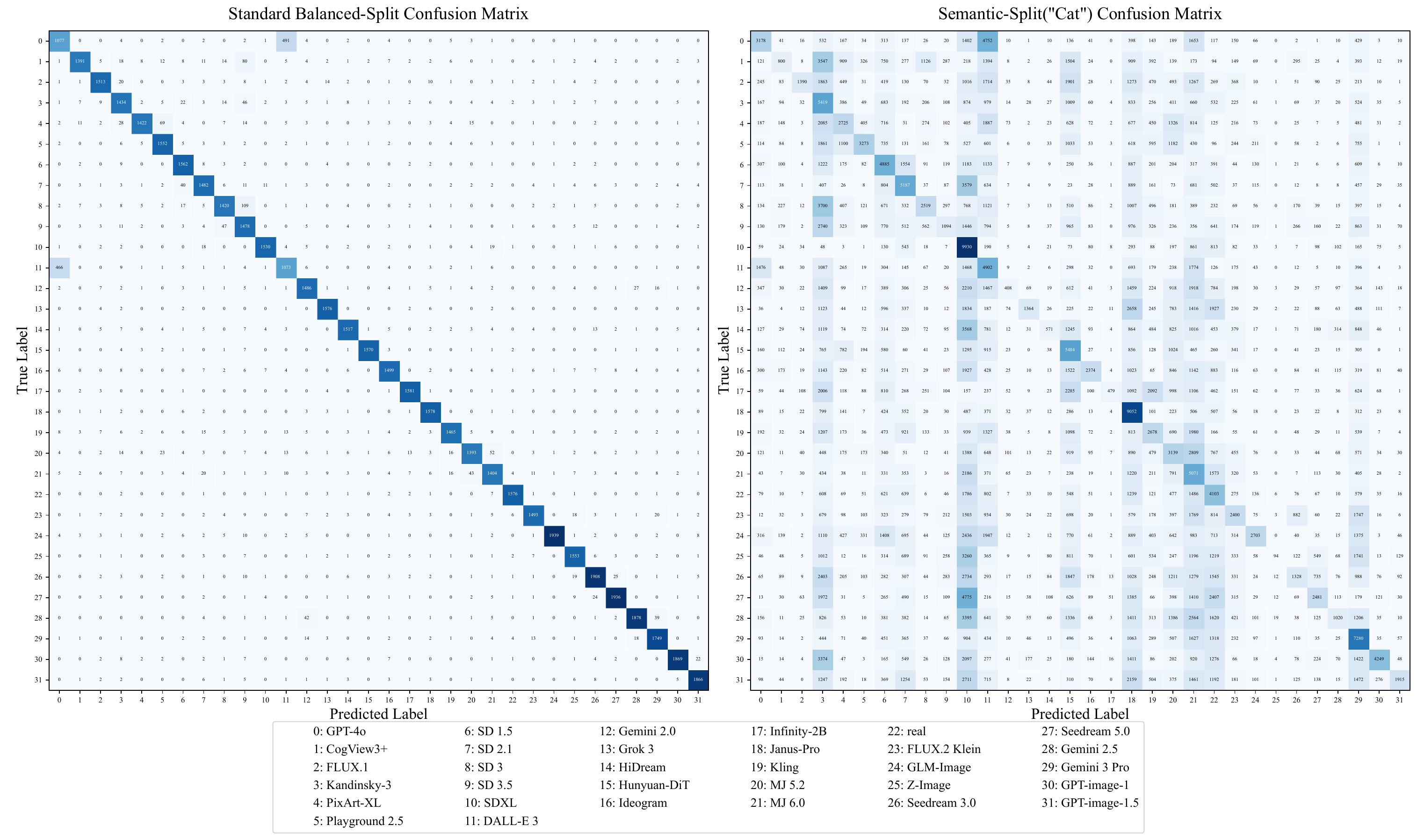}
  \caption{Confusion matrix of ResNet-50.}
  \label{fig:cm_resnet}
\end{figure*}

\begin{figure*}[h]
  \centering
  \includegraphics[width=\textwidth]{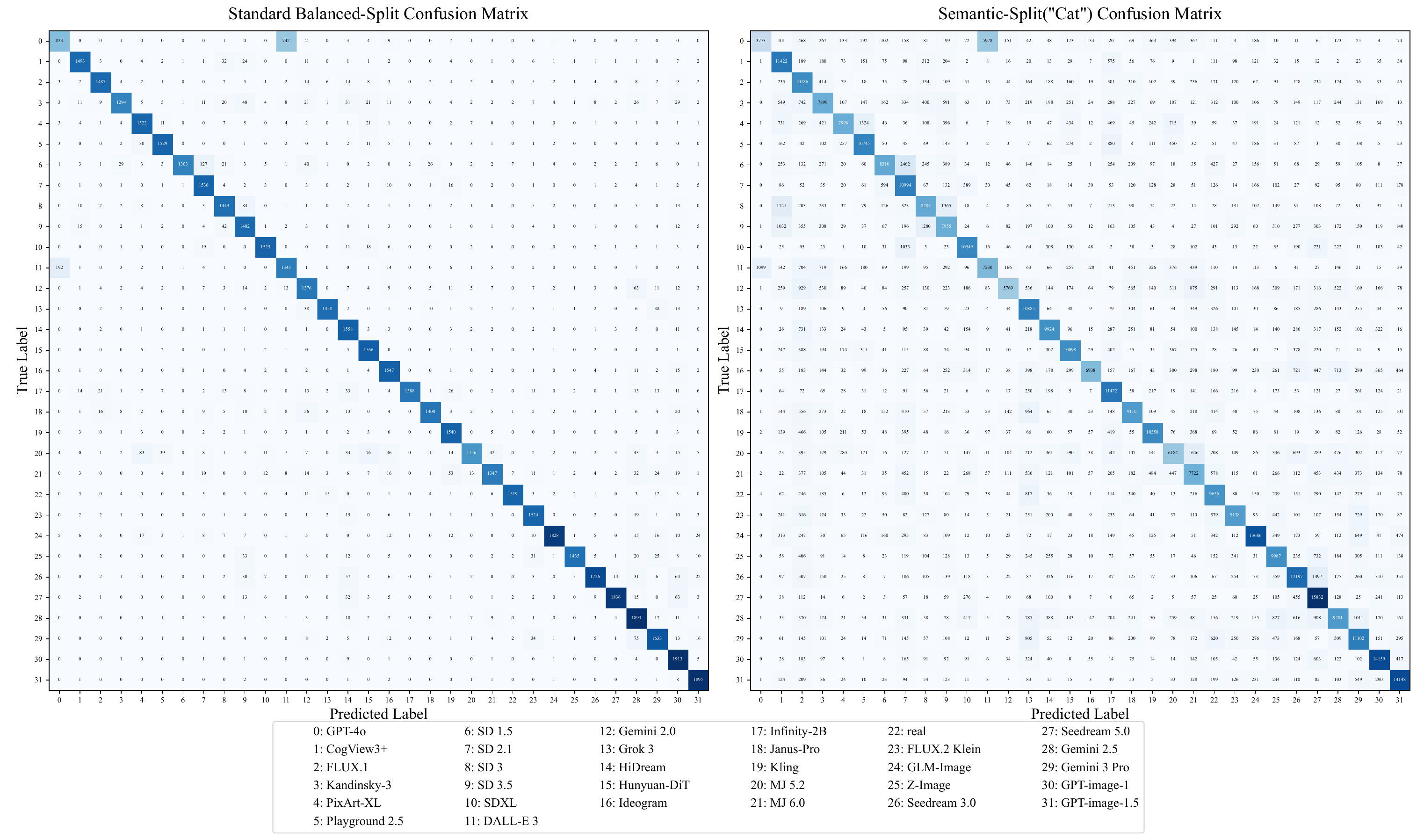}
  \caption{Confusion matrix of RepMix.}
  \label{fig:cm_repmix}
\end{figure*}

\paragraph{Confusion Matrix}
We present the confusion matrices for the baseline model ResNet-50 and the best-performing model RepMix, in Figure~\ref{fig:cm_resnet} and Figure~\ref{fig:cm_repmix}. While ResNet-50 demonstrates strong  performance in standard task, it exhibits relatively poor results in the semantic task, as indicated by the less prominent diagonal in the confusion matrix.  In contrast, RepMix displays clear diagonals in both matrices, highlighting its effectiveness in attribution. 

\begin{figure*}[h]
  \centering
  \includegraphics[width=\textwidth]{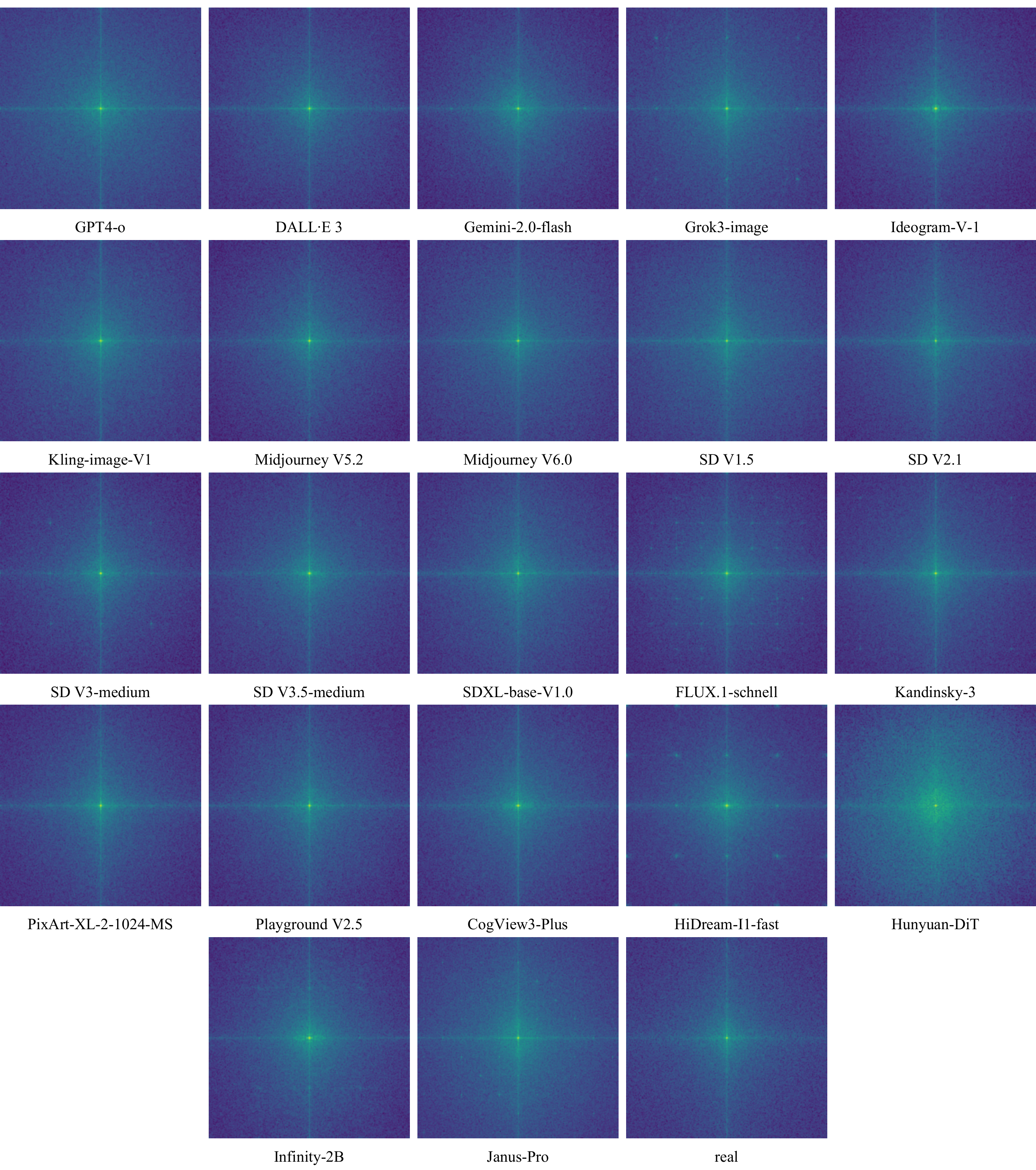}
\caption{Frequency spectra of generated images from a subset of generation sources.}
  \label{fig:freq}
\end{figure*}

\paragraph{Frequency Analysis}
Following \citet{corvi2023detection}, we extract noise residuals from 1,000 images per generation source and perform a Fourier transform on the average residual to generate the frequency spectrum, as shown in Figure~\ref{fig:freq}.  This figure demonstrates that most advanced generation models, including commercial models, cutting-edge DiTs and ARs leave fewer artifacts in the frequency domain compared to earlier GAN models. This indicates that the latest generators synthesize higher-quality images that are indistinguishable from real images, presenting a more challenging scenario for image attribution.


\section{Representative Image-Caption Pairs}
\begin{figure*}[h]
  \centering
  \includegraphics[width=0.8\textwidth]{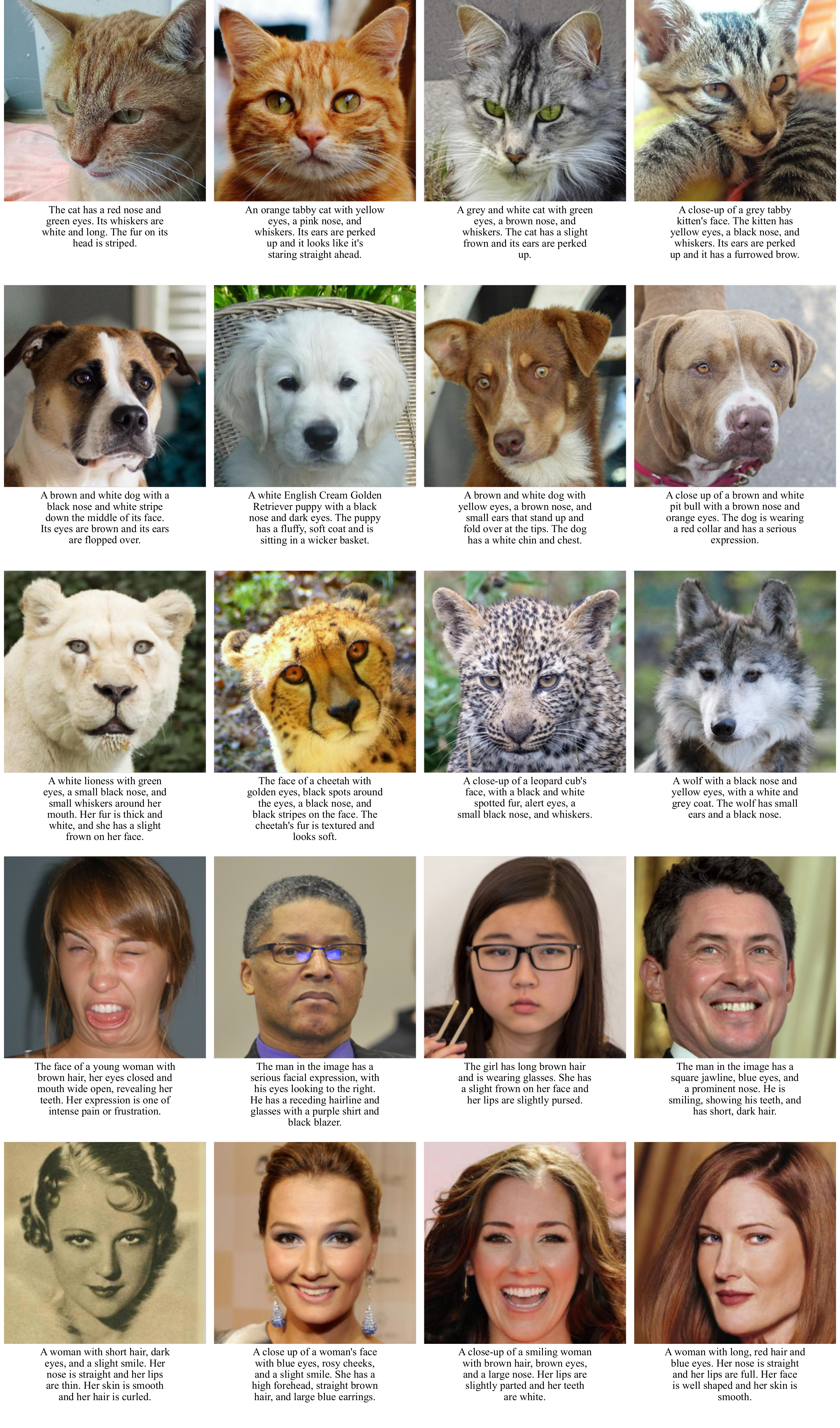}
  \caption{Representative image-caption pairs (upper half).}
  \label{fig:pair1}
\end{figure*}

\begin{figure*}[h]
  \centering
  \includegraphics[width=\textwidth]{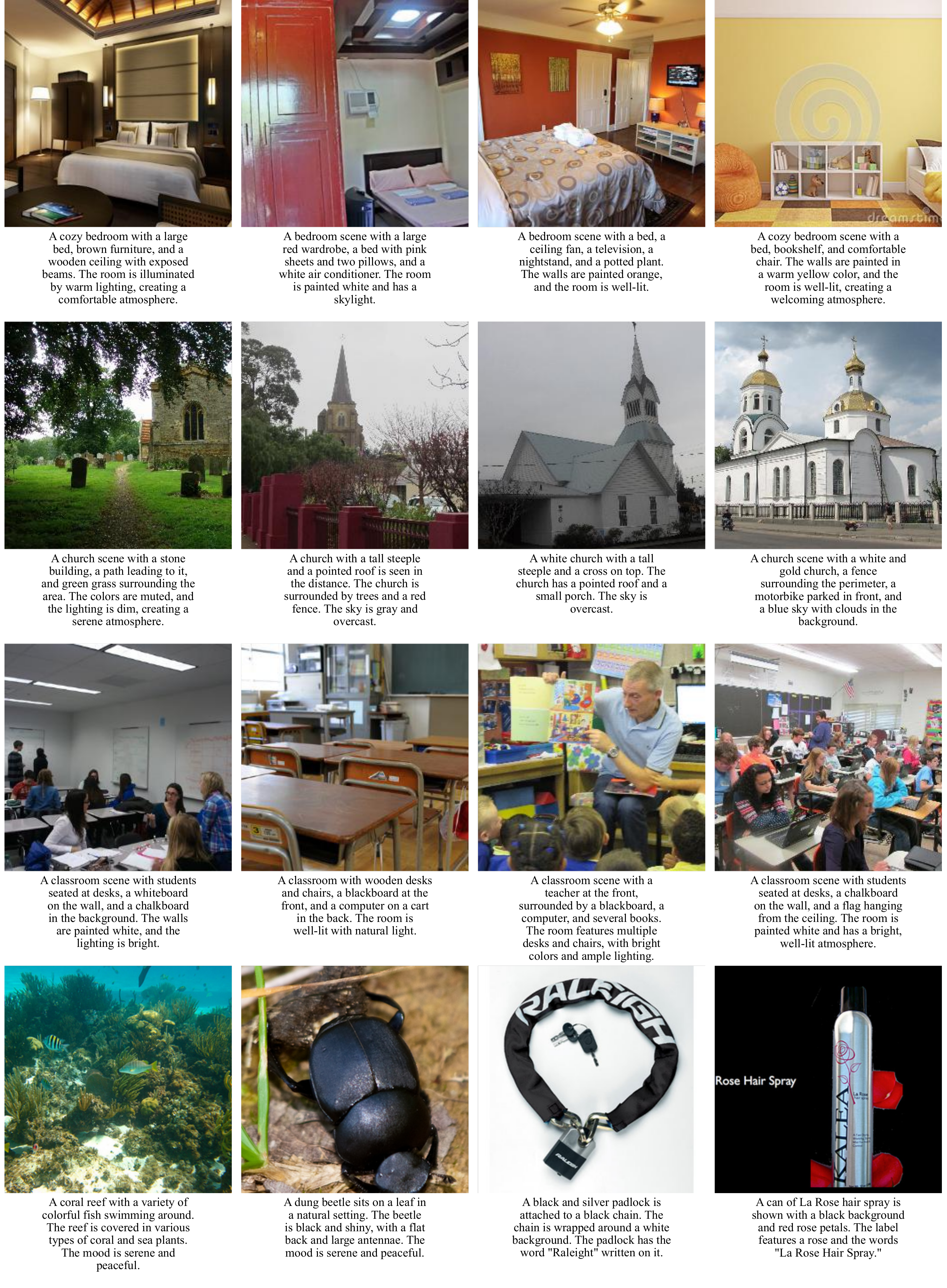}
  \caption{Representative image-caption pairs (lower half).}
  \label{fig:pair2}
\end{figure*}

In this section, we display some representative pairs of images and their corresponding captions generated by QwenVL-chat~\cite{Qwen-VL}, as illustrated in Figure~\ref{fig:pair1} and Figure~\ref{fig:pair2}. These figures demonstrate that the captions accurately describe the main objects in the images with precise details, thanks to the tailored prompt template and the exceptional captioning capabilities of QwenVL-chat. The precise captions enable text-to-image generation models to create images that closely adhere to the original in terms of semantic fidelity.

\section{Additional Visualization of Generated Images}
\label{Additional visualization}
Additional visualizations of generated images will be provided in the supplementary material instead of the appendix.

\begin{table*}[t]  
  \caption{Comparison of different methods training on ``Cat.''} 
  \label{tab:sp1}  
  \centering  
  \resizebox{\linewidth}{!}{
   \begin{tabular}{l c c c c c c c c}
    \toprule
    \multirow{3}{*}{Method}  
     & \multirow{3}{*}{\makecell[c]{Clean \\ Acc. (\%)}}  
     & \multicolumn{7}{c}{Degraded Acc.}  \\
    \cmidrule(lr){3-9}
    & 
     &  \makecell[c]{ DS \\ (0.5x)} 
     & \makecell[c]{ DS \\ (0.25x)} 
     & \makecell[c]{JPEG \\ (q=65)}  
     & \makecell[c]{JPEG \\ (q=30)}  
     & \makecell[c]{Blur \\ ($\sigma=3$)}  
     & \makecell[c]{Blur \\ ($\sigma=5$)}  
     & \makecell[c]{Avg \\ Acc.} \\
    \midrule
    ResNet-50~\cite{he2016deep} & 21.9 & 21.6 & 20.5 & 21.8 & 21.5 & 17.0 & 12.3 & 19.1 \\
    DCT-CNN~\cite{frank2020leveraging} & 23.2 & 19.9 & 15.3 & 18.9 & 16.8 & 7.8 & 3.3 & 13.7 \\
    HiFi-Net~\cite{guo2023hierarchical} & 50.8 & 46.6 & 29.5 & 17.5 & 12.7 & 32.1 & 10.4 & 24.8 \\
    \citet{li2024handcrafted} & 44.9 & 39.4 & 26.2 & 15.7 & 14.1 & 12.8 & 7.6 & 19.3 \\
    \midrule
    DNA-Det~\cite{yang2022deepfake} & \underline{68.6} & 60.3 & 27.2 & 13.6 & 11.4 & 30.0 & \textbf{28.5} & 28.5 \\
    RepMix~\cite{bui2022repmix} & 66.3 & \underline{65.1} & \textbf{59.2} & \textbf{65.6} & \textbf{63.1} & 30.4 & 18.2 & \textbf{50.3} \\
    \midrule
    PatchForensics~\cite{chai2020makes} & \textbf{69.1} & \textbf{66.6} & \underline{49.7} & 14.1 & 7.3 & \textbf{41.2} & 16.6 & 32.6 \\
    PatchCraft~\cite{zhong2023patchcraft} & 44.5 & 5.0 & 4.5 & 22.4 & 9.7 & 4.4 & 4.1 & 8.4 \\
    SSP~\cite{chen2024single} & 61.2 & 53.7 & 31.5 & 12.0 & 7.1 & 31.8 & 13.1 & 24.9 \\
    \midrule
    GFD~\cite{yang2021learning} & 45.9 & 45.0 & 40.8 & \underline{45.4} & \underline{41.9} & 20.4 & 10.5 & \underline{34.0} \\
    UCF~\cite{yan2023ucf} & 16.9 & 15.9 & 14.6 & 16.1 & 15.1 & 10.8 & 8.6 & 13.5 \\
        \midrule
FSD~\cite{wu2025fewshotlearnergeneralizesaigenerated} & 35.3 & 33.1 & 31.1 & 30.7 & 19.3 & 17.1 & 10.0 & 23.6 \\
    Omni-DFA~\cite{wu2025omnidfaunifiedframeworkopen} & 36.5 & 35.1 & 27.4 & 31.0 & 21.1 & \underline{32.4} & \underline{27.5} & 29.1 \\
    \midrule
    Avg & 45.0 & 39.0 & 29.0 & 25.0 & 20.1 & 22.2 & 11.6 & 24.7 \\
    \bottomrule  
  \end{tabular}  
  }
\end{table*}

\begin{table*}[t]  
  \caption{Comparison of different methods training on ``Bedroom.''} 
  \label{tab:sp2}  
  \centering  
  \resizebox{\linewidth}{!}{
   \begin{tabular}{l c c c c c c c c}
    \toprule
    \multirow{3}{*}{Method}  
     & \multirow{3}{*}{\makecell[c]{Clean \\ Acc. (\%)}}  
     & \multicolumn{7}{c}{Degraded Acc.}  \\
    \cmidrule(lr){3-9}
    & 
     &  \makecell[c]{ DS \\ (0.5x)} 
     & \makecell[c]{ DS \\ (0.25x)} 
     & \makecell[c]{JPEG \\ (q=65)}  
     & \makecell[c]{JPEG \\ (q=30)}  
     & \makecell[c]{Blur \\ ($\sigma=3$)}  
     & \makecell[c]{Blur \\ ($\sigma=5$)}  
     & \makecell[c]{Avg \\ Acc.} \\
    \midrule
    ResNet-50~\cite{he2016deep} & 18.0 & 18.0 & 16.4 & 17.3 & 16.7 & 10.1 & 9.0 & 14.6 \\
    DCT-CNN~\cite{frank2020leveraging} & 30.9 & 26.8 & 17.8 & 19.8 & 17.5 & 7.3 & 1.9 & 15.2 \\
    HiFi-Net~\cite{guo2023hierarchical} & 61.9 & 43.6 & 30.4 & 28.3 & 20.0 & 14.6 & 7.0 & 24.0 \\
    \citet{li2024handcrafted} & 43.2 & 37.6 & 23.0 & 15.1 & 12.4 & 10.3 & 7.0 & 17.6 \\
    \midrule
    DNA-Det~\cite{yang2022deepfake} & \underline{70.6} & 63.5 & 41.1 & 12.5 & 13.0 & 24.9 & 7.5 & 27.1 \\
    RepMix~\cite{bui2022repmix} & 66.6 & \underline{64.8} & \textbf{58.9} & \textbf{63.9} & \textbf{60.9} & 24.9 & \underline{19.2} & \textbf{48.8} \\
    \midrule
    PatchForensics~\cite{chai2020makes} & \textbf{73.4} & \textbf{71.1} & \underline{53.8} & 14.5 & 7.8 & \textbf{44.9} & 18.4 & \underline{35.1} \\
    PatchCraft~\cite{zhong2023patchcraft} & 39.1 & 6.0 & 3.7 & 14.5 & 9.0 & 5.2 & 4.3 & 7.1 \\
    SSP~\cite{chen2024single} & 61.3 & 50.8 & 32.8 & 12.4 & 10.1 & 29.4 & 9.5 & 24.2 \\
    \midrule
    GFD~\cite{yang2021learning} & 43.0 & 37.5 & 31.0 & \underline{39.9} & \underline{35.8} & 6.3 & 3.7 & 25.7 \\
    UCF~\cite{yan2023ucf} & 17.9 & 16.4 & 14.9 & 15.3 & 13.7 & 8.1 & 6.0 & 12.4 \\
    \midrule
    FSD~\cite{wu2025fewshotlearnergeneralizesaigenerated} & 34.3 & 31.1 & 20.7 & 31.5 & 14.1 & 15.3 & 9.3 & 20.3 \\
    Omni-DFA~\cite{wu2025omnidfaunifiedframeworkopen} & 39.5 & 37.6 & 25.9 & 31.5 & 19.8 & \underline{34.5} & \textbf{28.4} & 29.6 \\
    \midrule
    Avg & 46.7 & 38.9 & 28.7 & 23.7 & 19.3 & 16.8 & 8.6 & 23.2 \\
    \bottomrule  
  \end{tabular}  
  }
\end{table*}

\begin{table*}[t]  
  \caption{Comparison of different methods training on ``Human Face.''} 
  \label{tab:sp3}  
  \centering  
  \resizebox{\linewidth}{!}{
   \begin{tabular}{l c c c c c c c c}
    \toprule
    \multirow{3}{*}{Method}  
     & \multirow{3}{*}{\makecell[c]{Clean \\ Acc. (\%)}}  
     & \multicolumn{7}{c}{Degraded Acc.}  \\
    \cmidrule(lr){3-9}
    & 
     &  \makecell[c]{ DS \\ (0.5x)} 
     & \makecell[c]{ DS \\ (0.25x)} 
     & \makecell[c]{JPEG \\ (q=65)}  
     & \makecell[c]{JPEG \\ (q=30)}  
     & \makecell[c]{Blur \\ ($\sigma=3$)}  
     & \makecell[c]{Blur \\ ($\sigma=5$)}  
     & \makecell[c]{Avg \\ Acc.} \\
    \midrule
    ResNet-50~\cite{he2016deep} & 26.3 & 25.9 & 24.0 & 26.0 & 25.5 & 22.6 & 16.8 & 23.5 \\
    DCT-CNN~\cite{frank2020leveraging} & 34.0 & 30.7 & 20.6 & 20.9 & 16.1 & 5.6 & 1.4 & 15.9 \\
    HiFi-Net~\cite{guo2023hierarchical} & 56.4 & 47.4 & 29.6 & 27.5 & 18.6 & 22.4 & 8.2 & 25.6 \\
    \citet{li2024handcrafted} & 51.1 & 43.2 & 25.4 & 15.4 & 13.8 & 13.5 & 9.7 & 20.2 \\
    \midrule
    DNA-Det~\cite{yang2022deepfake} & \textbf{74.6} & \underline{62.8} & 30.7 & 10.2 & 8.3 & 37.4 & 9.1 & 26.4 \\
    RepMix~\cite{bui2022repmix} & 59.9 & 58.2 & \textbf{53.7} & \textbf{63.8} & \textbf{60.5} & 34.5 & \underline{24.0} & \textbf{49.1} \\
    \midrule
    PatchForensics~\cite{chai2020makes} & \underline{67.2} & \textbf{64.3} & \underline{42.0} & 9.5 & 5.4 & \textbf{43.3} & 13.9 & 29.7 \\
    PatchCraft~\cite{zhong2023patchcraft} & 39.0 & 5.3 & 5.4 & 16.5 & 11.6 & 5.0 & 4.9 & 8.1 \\
    SSP~\cite{chen2024single} & 62.0 & 54.0 & 28.9 & 10.1 & 7.1 & 30.2 & 10.8 & 23.5 \\
    \midrule
    GFD~\cite{yang2021learning} & 25.7 & 22.7 & 22.7 & 29.5 & 31.1 & 29.2 & 13.1 & 22.3 \\
    UCF~\cite{yan2023ucf} & 21.8 & 21.4 & 19.5 & 21.1 & 19.8 & 13.8 & 9.7 & 17.6 \\
    \midrule
    FSD~\cite{wu2025fewshotlearnergeneralizesaigenerated} & 36.8 & 35.3 & 26.2 & 25.6 & 15.4 & 15.9 & 9.8 & 21.4 \\
    Omni-DFA~\cite{wu2025omnidfaunifiedframeworkopen} & 43.4 & 41.5 & 28.2 & \underline{39.1} & \underline{32.8} & \underline{38.9} & \textbf{33.5} & \underline{35.7} \\
    \midrule
    Avg & 45.8 & 39.4 & 28.0 & 24.4 & 20.3 & 22.8 & 12.3 & 24.5 \\
    \bottomrule  
  \end{tabular}  
  }
\end{table*}


\clearpage
\clearpage

\clearpage
\newpage
\newpage
\newpage
\newpage



\clearpage

\end{document}